\def\year{2020}\relax
\documentclass[letterpaper]{article} 
\usepackage{aaai20}  
\usepackage{times}  
\usepackage{helvet} 
\usepackage{courier}  
\usepackage[hyphens]{url}  
\usepackage{graphicx} 
\usepackage{graphicx}  
\usepackage{amsfonts}
\usepackage{amsmath}
\usepackage{microtype}

\newcommand{\subsubsubsection}[1]{\noindent\textbf{#1.}}
\newcommand{\ie}{i.e.}

\newcommand{\eg}{eg.}
\newcommand{\Eg}{Eg.}
\newcommand{\wrt}{w.r.t.}
\newcommand{\model}{DeepDualMapper\xspace} 

\newcommand{\citenp}[1]{\citeauthor{#1} \citeyear{#1}}

\usepackage{xspace}
\usepackage{subfigure}

\title{DeepDualMapper: A Gated Fusion Network for Automatic Map Extraction using Aerial Images and Trajectories}

\author{Hao Wu\textsuperscript{1,2,3,5,\thanks{indicates equal contribution.}}, Hanyuan Zhang\textsuperscript{1,2,3,5,\footnotemark[1]}, Xinyu Zhang\textsuperscript{1,2,3,5}, Weiwei Sun\textsuperscript{1,2,3}, \\\bf \Large Baihua Zheng\textsuperscript{4}, Yuning Jiang\textsuperscript{5}\\
	\\
	\textsuperscript{1}School of Computer Science, Fudan University\\
	\textsuperscript{2}Systems and Shanghai Key Laboratory of Data Science, Fudan University\\
	\textsuperscript{3}Shanghai Insitute of Intelligent Electroics \& Systems\\
	\textsuperscript{4}Singapore Management University, \textsuperscript{5}Bytedance AI Lab\\
	{\small \{wuhao5688, hanyuanzhang16, zhangxinyu, wwsun\}@fudan.edu.cn}\\
	{\small bhzheng@smg.edu.sg, jiangyuning@bytedance.com}
}

\begin{document}
	
	\maketitle
	
	\begin{abstract}
		Automatic map extraction is of great importance to urban computing and location-based services. 
		Aerial image and GPS trajectory data refer to two different data sources that could be leveraged to generate the map, although they carry different types of information.
		Most previous works on data fusion between aerial images and data from auxiliary sensors do not fully utilize the information of both modalities and hence suffer from the issue of information loss. 
		%
		%
		We propose a deep convolutional neural network called \model which fuses the aerial image and trajectory data in a more seamless manner 
		to extract the digital map. We design a gated fusion module to explicitly control the information flows from both modalities in a complementary-aware manner. Moreover, we propose a novel densely supervised refinement decoder to generate the prediction in a coarse-to-fine way. 
		Our comprehensive experiments demonstrate that \model can fuse the information of images and trajectories much more effectively than existing approaches, and is able to generate maps with higher accuracy. 
	\end{abstract}
	
	\section{Introduction}

Generating city maps is a fundamental building block of many location-based applications like navigation, autonomous vehicle, and so on. 
Multiple data sources can be leveraged to automatically generate the map. In this paper, we focus on two data sources, namely \emph{aerial images} and \emph{GPS trajectories}. 
The former is captured by the satellites. 
%
Such images are publicly available in many places, \eg, Google Earth can stay up-to-date, and have been explored extensively in the past decades for map extraction
(
\citenp{roadtracer}; \citenp{dlinknet}; \citenp{stackedunet}; \citenp{robocodes}, \citenp{hinton_mapinference}; \citenp{casnet}).
The latter captures human movements in the urban area as most human movements are constrained by the underlying road network. 
Thanks to the fast development of mobile platforms and localization techniques, collecting these GPS trajectories is much easier nowadays. How to extract the map from trajectory data is also a hot research problem (\citenp{kde}; \citenp{tc1}; 
\citenp{trajsift}; \citenp{cobweb}).

\begin{figure}[t]
	\centering
	\includegraphics[width=70mm]{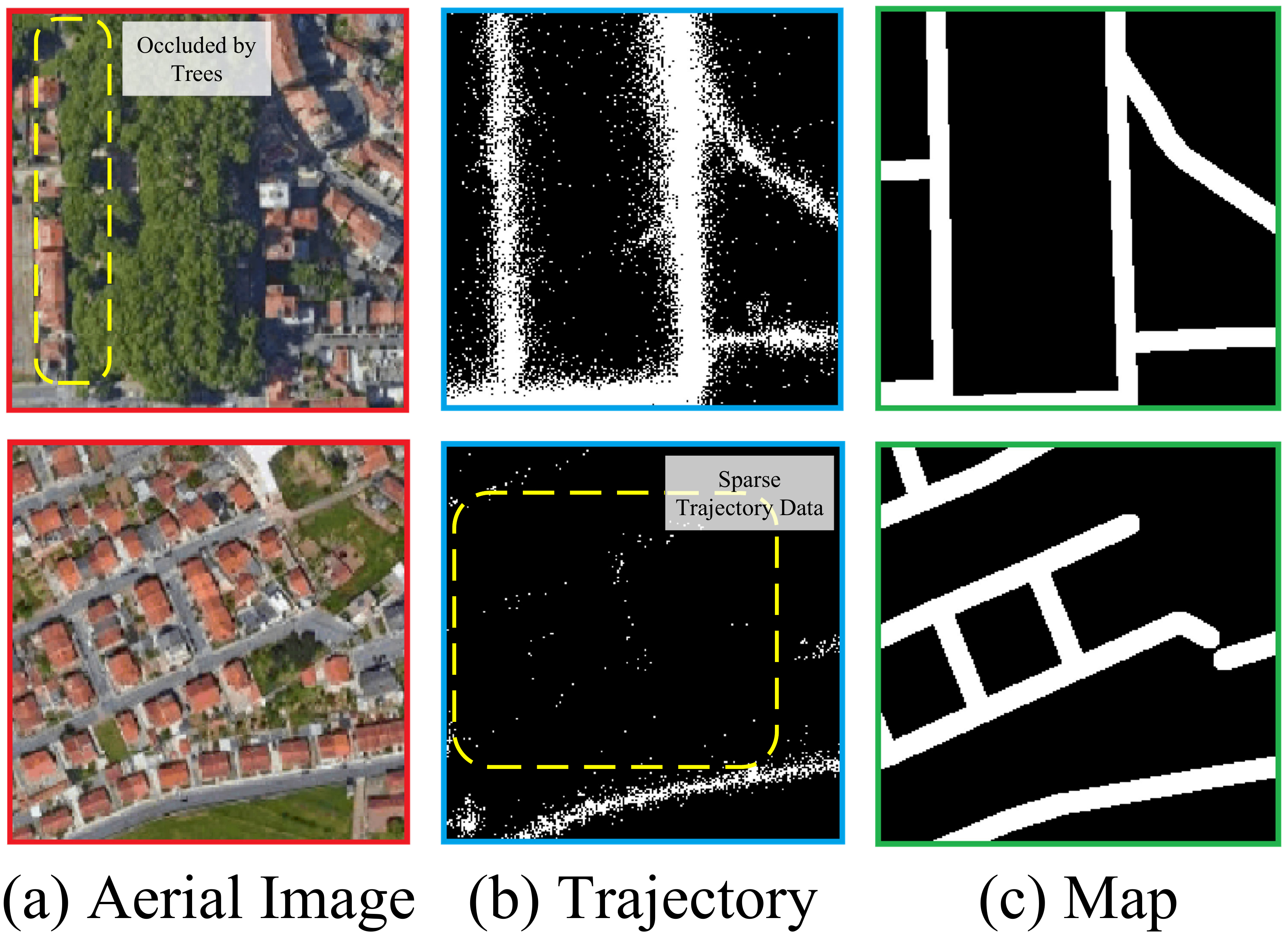}
	\caption{Examples showing some limitations in aerial images and trajectories (visualized by plotting the GPS points). \Eg, roads are covered by trees and trajectories are sparse on minor roads. However, by leveraging both data sources, such problems are expected to be eliminated.}
	\label{fig:data_flaw}
\end{figure}

However, neither data source is perfect for the task of automatic map extraction because of the \emph{information loss} issue. 
There is inevitably missing information in the data which makes it extremely challenging, if not impossible, to infer the map accurately. 
\Eg, as shown in Fig.~\ref{fig:data_flaw}, in aerial image, some roads may be occluded by trees and buildings; in trajectory data, some less popular roads may have very few, or zero, historical trajectories passing by. Such information loss introduces more complexity to the map extraction process.

As aerial images and trajectory data capture different types of information, the information missed in one source might be available in the other source. 
In other words, combining two data sources that are complementary offers an effective way to make full use of the information that can be well captured by at least one data source. In the current literature, a variety of researches have tried to fuse aerial images with other auxiliary data sources, including earth observation data including Radar \cite{multi3net}, Lidar (\citenp{beyondrgb}; \citenp{l3fsn}; \citenp{triseg}), OpenStreetMap data \cite{rgbosmfuse} and street view images \cite{anotherfusenet}, to solve the urban scene semantic segmentation problem, which, to some extent, is related to map extraction task. 
However, these approaches either simply concatenate the features of two modalities  (i.e., early-fusion), or compute the average of the predictions made by each modality (i.e., late-fusion). We argue that such rough designs fail to fully explore the utilities of both modalities and are not able to effectively address the issue of information loss faced by both modalities. 
Among these approaches, the most relevant work solves the same problem by proposing a new deconvolution strategy called \emph{1D decoder}~\cite{1ddecoder}. However, that work also simply concatenates the image features and the trajectory heat map which leaves room for further improvement.

Consequently, we propose a novel fusion model, namely \model, which aims to fuse the aerial image and trajectory data more seamlessly. 
We design a \emph{Gated Fusion Module} (GFM) to explicitly learn the modality selection based on the confidence of each data source. 
It controls the information flows of aerial images and trajectories by \emph{explicitly} defining a learnable weight to make the fusion process complementary-aware, \ie, giving a higher weight to the data source that is more trustworthy and more valuable to infer the answer. Such a design is consistent with the human decision-making process when we need to choose one between two data sources with different confidences. 
We also introduce a refinement process, conducted through the \emph{Densely Supervised Refinement} (DSR) strategy, to refine the prediction generated by GFM from coarse to fine via residual refinement learning. 

In summary, we have made three major contributions in this paper. 
First, we propose a novel data fusion model called \model which utilizes both aerial images and trajectory data more effectively for the task of map extraction. 
Second, we design a novel gated fusion module and a refinement decoder that can adaptively control the information flow of both modalities and select the one being more reliable in a coarse-to-fine refinement manner. Such a design follows the heuristics of the human decision-making process when judging two data sources.
Third, our model not only outperforms the baselines in three real datasets but also demonstrates a superior resilience to information loss as it can generate the map with higher accuracy than all the existing competitors even when both modalities have 25\% information loss. 

\section{Related work}

\paragraph{Map Extraction via Aerial Images.}
Many approaches have been proposed to extract a digital map from aerial images.
\cite{hinton_mapinference} uses a feed forward neural network to detect the roads with unsupervised pre-training.
After CNN has demonstrated its powerful predictability on visual tasks, many approaches adopt CNNs on this task, such as \cite{roadandbuildingcnn} (with FCN), \cite{casnet} (with DeconvNet), and \cite{robocodes} (with SegNet).
\cite{deeproadmapper} and \cite{roadtracer} are two representative approaches proposed recently for extracting road network topology instead of directly extracting a binary representation of the map. \cite{dlinknet} and \cite{stackedunet} adopt U-Net-like structures to generate the maps.
These approaches purely rely on the aerial image which has its limitations (\eg roads covered by trees and shadows are not visible) and require high-quality optical sensors.

\paragraph{Map Extraction via Trajectory Data.}
Approaches leveraging trajectory data mainly rely on clustering techniques to cluster the trajectories or GPS points located on the same road and then to extract a road from each cluster.  
%
\cite{kmeans_mapinference} performs k-means style clustering on GPS samples to generate road. 
\cite{tracemerge} clusters the trajectories into a growing road network. 
\cite{kde} is a representative method transforming trajectories into a discretized image and adopts kernel density estimation to extract the road network.
Among these approaches, \cite{tc1} is reported as the one achieving the best performance \cite{mapinference_eval}.
Recently, \emph{map update} has also attracted some attentions, which is to incrementally discover new roads and add them to the existing road network (\citenp{crowdatlas}; \citenp{cobweb}; \citenp{hymu}; \citenp{glue}; \citenp{trajsift}).
%
In general, these trajectory-based approaches solve the problem in heuristics but not learning-based manner, which makes them inevitably inferior to those learning-based approaches (will be shown in experiments).
 
\paragraph{Data Fusion for Aerial Images.}
In the literature, there are some research works using aerial images with the assistance of other remote sensing data sources to reinforce the result. 
Semantic labeling of urban areas is a representative application for fusion of aerial images and other sensors such as the Lidar point cloud sensor data (\citenp{beyondrgb}; \citenp{l3fsn}), laser data \cite{segnetrc}, OpenStreetMap data \cite{rgbosmfuse} and street view images \cite{anotherfusenet}. Late fusion strategy is adopted by (\citenp{rgbosmfuse}; \citenp{segnetrc}; \citenp{beyondrgb}; \citenp{l3fsn}). \cite{segnetrc} improves it by a residual correction strategy called SegNet-RC. Encoder fusion strategy is applied in (\citenp{rgbosmfuse}; \citenp{beyondrgb}; \citenp{anotherfusenet}; \citenp{l3fsn}) via FuseNet \cite{fusenet}. V-FuseNet, an advanced version, is proposed in \cite{beyondrgb}. 
Besides, \cite{multi3net} uses multiple sensors including satellite image to segment the flooded buildings. It fuses multiples sensors through late fusion strategy.
\cite{triseg} and \cite{1ddecoder} are the only existing works for fusing aerial images with other auxiliary data source to detect the roads. The former fuses the RGB image with Lidar data while the latter fuses the image with trajectory data to generate the map. 
However, most of these works either simply concatenate the features or average the predictions of both modalities to perform data fusion and hence have not fully utilized both data sources. 

\begin{figure*}[t]
	\centering
	\includegraphics[width=180mm]{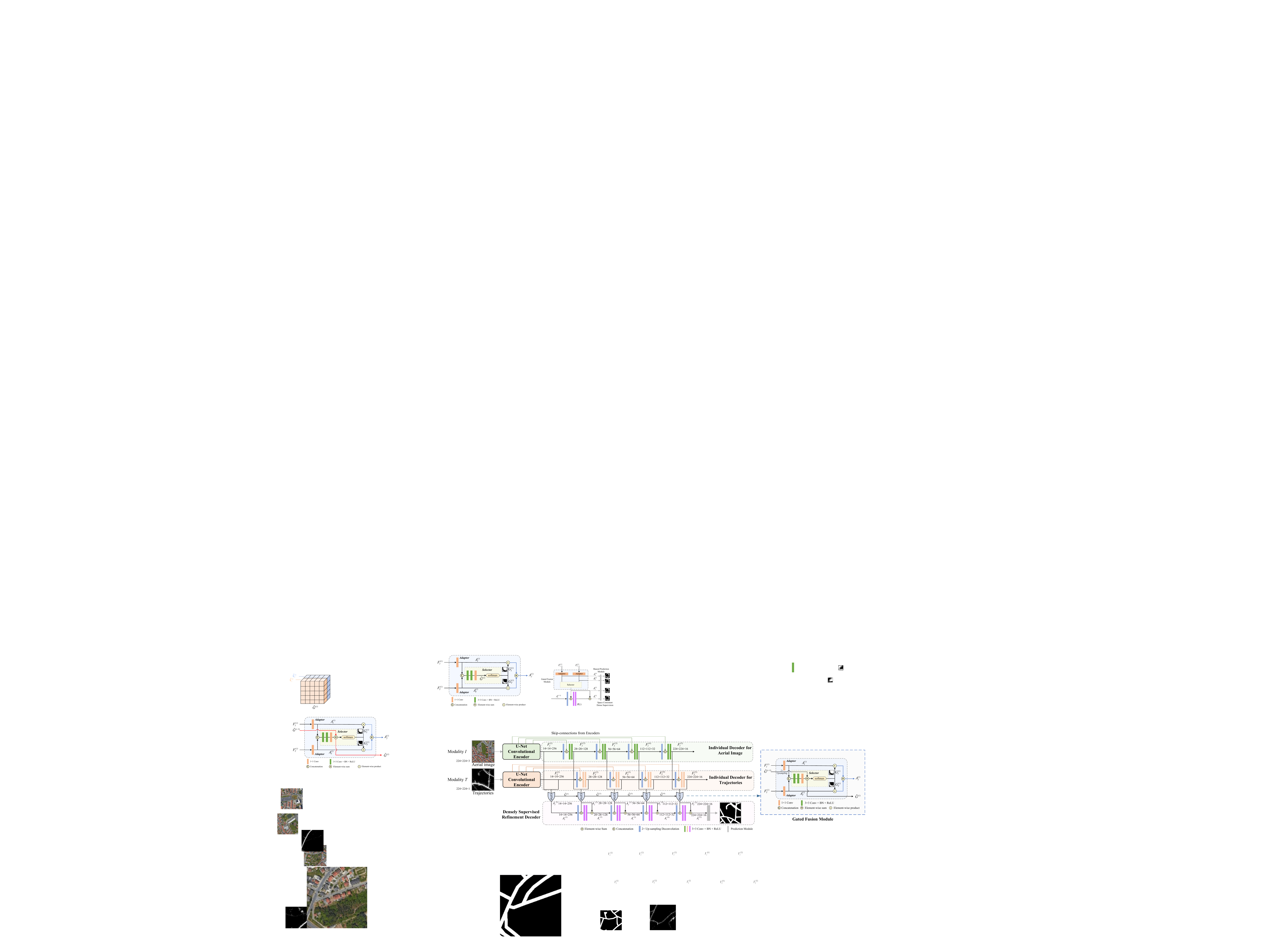}
	\caption{The main fusion architecture of our model. The right part illustrates the details of the gated fusion module. We omit the details of the encoder as it is the same as U-Net. Note that all dense supervisions do not appear in the figure due to the clarity of the image. Please refer to Fig.~\ref{fig:refinement} for the details. All annotated dimensions are assumed by feeding the input in the size of $\mathbf{224 \times 224}$, for a clearer understanding of each layer. Best viewed in color.}
	\label{fig:model}
\end{figure*}

\section{Deep Dual Mapper}

\subsection{Model Overview}
We follow the flow of image-based map extraction approaches to transfer the map extraction task into a pixel-wise binary prediction task (\citenp{dlinknet}; \citenp{stackedunet}; \citenp{casnet}; \citenp{1ddecoder}). We select U-Net as the main framework for its excellent performance in semantic segmentation, especially biomedical image segmentation, with a simple and elegant structure \cite{unet}. For the input data, we crop the whole area by $224 \times 224$ grids and the trajectory feature is represented by counting the historical GPS points in each pixel. Please refer to the supplementary material for more details on the model and data preprocessing procedure. 

Fig.~\ref{fig:model} illustrates the main architecture of \model. It consists of two independent U-Nets with each taking in a branch of data source as the input. The original decoders of those two U-Nets serve as the auxiliary decoders to preserve the key information of aerial images and that of trajectories respectively. We extract the activation maps of two branches \wrt all scales, \ie, $F_I^{(1)} \sim F_I^{(5)}$ and $F_T^{(1)} \sim F_T^{(5)}$ and fuse them through the \emph{gated fusion module} which produces the fused information $A_f^{(1)} \sim A_f^{(5)}$. Then, we propose the main decoder, the \emph{densely supervised refinement decoder}, which shares the same structure as U-Net's decoder. It takes the fused features $A_f^{(i)}$ as input and generates the pixel-wise prediction. Finally, it generates a feature map that shares the same spatial dimension as the original input dimension and adopts a linear binary predictor to generate pixel-wise predictions. In the following sections, we will detail the gated fusion module and the densely supervised refinement decoder.

\subsection{Gated Fusion Module}\label{sec:gfm}
One of the key novelties of our model is the Gated Fusion Module (GFM), which is proposed to simulate the human decision-making process. Given an analytic task and two data sources, we normally select the data source that is more informative and provides more valuable input for the given task. That is to say, for the task of map extraction, we prefer the data source that makes the inference of the roads much easier and provides more useful information to the other one. 
%
To be more specific, it takes the activation maps from two modalities as inputs, \ie, $F_I^{(i)}$ for aerial image and $F_T^{(i)}$ for trajectory where $i$ refers to the level (1 $\sim$ 5), and outputs the fused features $A_{f}^{(i)} = GFM^{(i)}\left(F_I^{(i)}, F_T^{(i)}\right)$.
GFM is composed of two sub-modules, \ie, adapter and selector, as detailed in the following. 

As the two activation maps are extracted from two independent branches and there is a potential space inconsistency between two modalities, directly fusing them is not an ideal solution. Accordingly, the adapters are introduced to transfer the activation spaces $\mathcal{F}_I$ and $\mathcal{F}_T$ to a uniform space $\mathcal{A}$ to enable a linear combination of the two activation maps in such a uniform space. 
We denote the adapted features as $A_I^{(i)}= a_I\left(F_I^{(i)}\right) \in {\mathbb R}^{h^{(i)}\times w^{(i)} \times c^{(i)}}$ and $A_T^{(i)} = a_T\left(F_T^{(i)}\right)\in {\mathbb R}^{h^{(i)}\times w^{(i)} \times c^{(i)}}$. Here, $a_I(\cdot)$ and $a_T(\cdot)$ are the channel dimension-preserving adapter operations ${\mathbb R}^{c^{(i)}} \longmapsto {\mathbb R}^{c^{(i)}}$ implemented by the $1\times 1$ convolution, i.e., the channel dimension is linearly transferred into a uniform space $\mathcal{A}$. $h(i)$, $w(i)$, and $c(i)$ refer to the height, width and channel dimension of the feature map respectively. The activation maps are in a uniform space after the transformation, and they can be safely fused by a linear combination, \ie, 
\[A_f^{(i)} = G_I^{(i)} \odot A_I^{(i)} + G_T^{(i)} \odot A_T^{(i)},~~~s.t.,~~G_I^{(i)} + G_T^{(i)} = \textbf{1},\]
where $G_I^{(i)} \in {\mathbb R}^{h^{(i)}\times w^{(i)}}$ and $G_T^{(i)} \in {\mathbb R}^{h^{(i)}\times w^{(i)}}$ refer to the gate values \wrt~aerial image and trajectory respectively; $\odot$ denotes the element-wise product. The complementary constraint $G_I^{(i)} + G_T^{(i)} = \textbf{1}$ enforces the module to learn a complementary-style fusion. It implies that if certain area in one modality has higher confidence in  generating the prediction, it will be assigned a larger weight which simultaneously reduces the weight of the other modality with lower confidence in prediction. If both modalities have sufficient useful information to make a good prediction, the exact values of their weights are not important.

\begin{figure}[t]
	\centering
	\includegraphics[width=65mm]{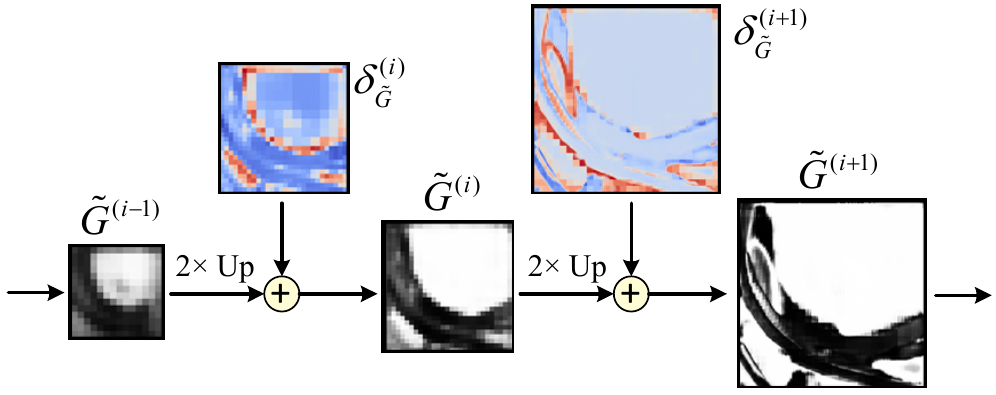}
	\caption{An example showing the coarse-to-fine gate refinement procedure.} 
	\label{fig:gate_res}
\end{figure}

The selector sub-module computes the gate values $G_I^{(i)}$ and $G_T^{(i)}$ based on the information of two modalities, \ie, $A_I^{(i)}$ and $A_T^{(i)}$. To get the values of $G_I^{(i)}$ and $G_T^{(i)}$, we first compute the un-normalized predictions $\tilde G^{(i)}  \in {\mathbb R}^{h^{(i)}\times w^{(i)} \times 2}$. $\tilde G_I^{(i)}=\tilde G^{(i)}[:,:,0]$ and $\tilde G_T^{(i)}=\tilde G^{(i)}[:,:,1]$ are extracted from the slices of $\tilde G^{(i)}$. Softmax function is adopted to normalize the gate value \wrt~each location in $h^{(i)} \times w^{(i)}$ to meet the complementary constraint $G_I^{(i)} + G_T^{(i)} = \textbf{1}$ as shown in Eq.~(\ref{eq:softmax}). The un-normalized gates $\tilde G^{(i)}$ is computed in a recursive manner formulated in Eq.~(\ref{eq:gate_refinement}), which means the decision of the gate at level $i$ is based on the decision made at level ($i-1$) with some residual refinement $\delta^{(i)}_{\tilde G}$. The intuition is that we want to perform a consistent regularization on the gate decision space across all scales, i.e., the gates generated from different resolution granularities should not differ much, as the example shown in Fig.~\ref{fig:gate_res}. Note that $\tilde G^{(i-1)}$ has its scale different from $\tilde G^{(i)}$, which can be solved by a $2\times$ non-parametric up-sampling operation $U_{2\times}(\cdot)$. For the residual refinement $\delta^{(i)}_{\tilde G}$, it is fed by the concatenation of $A_I^{(i)}$ and $A_T^{(i)}$ (i.e., $\textcircled{c}$ denotes the concatenation of two feature maps along the channel dimension), followed by two $3\times 3$ convolution blocks, denoted by $\phi^2_{3\times3}$, with batch normalization and ReLU activation. We adopt the stack of $3\times 3$ convolutions in order to enlarge the receptive field and to collect more global information from those two modalities. Hence, it can make more accurate decisions by a pixel-wise linear transformation which is implemented by an $1\times 1$ convolution $\psi_{1\times 1}$ with the output channel at 2. The whole computation flow of the GFM is illustrated in the right part of Fig.~\ref{fig:model}.
\begin{align}\notag
\delta^{(i)}_{\tilde G}  &= \psi_{1\times1}\left(\phi^2_{3\times3}\left(A_I^{(i)} \textcircled{c}~ A_T^{(i)}\right)\right) \in {\mathbb R}^{h^{(i)}\times w^{(i)} \times 2}\\\label{eq:gate_refinement}
{\tilde G^{(i)}} &= U_{2\times}\left(\tilde G^{(i-1)}\right) + \delta^{(i)}_{\tilde G} \in {\mathbb R}^{h^{(i)}\times w^{(i)} \times 2} \\\label{eq:softmax}
G_I^{(i)} &= \frac{\exp{\tilde G_I^{(i)}}}{\exp{\tilde G_I^{(i)}} + \exp{\tilde G_T^{(i)}}}, G_T^{(i)} = \frac{\exp{\tilde G_T^{(i)}}}{\exp{\tilde G_I^{(i)}} + \exp{\tilde G_T^{(i)}}}
\end{align}

\subsection{Densely Supervised Refinement Decoding}
The GFM aims to simulate the human decision-making process for judging the information from two modalities. In this section, we propose a densely supervised refinement decoder (DSRD) to further process the fused information into the prediction map. We introduce it through following two operations, i.e., the fusion refinement and the dense supervision.

\begin{figure}[t]
	\centering
	\includegraphics[width=70mm]{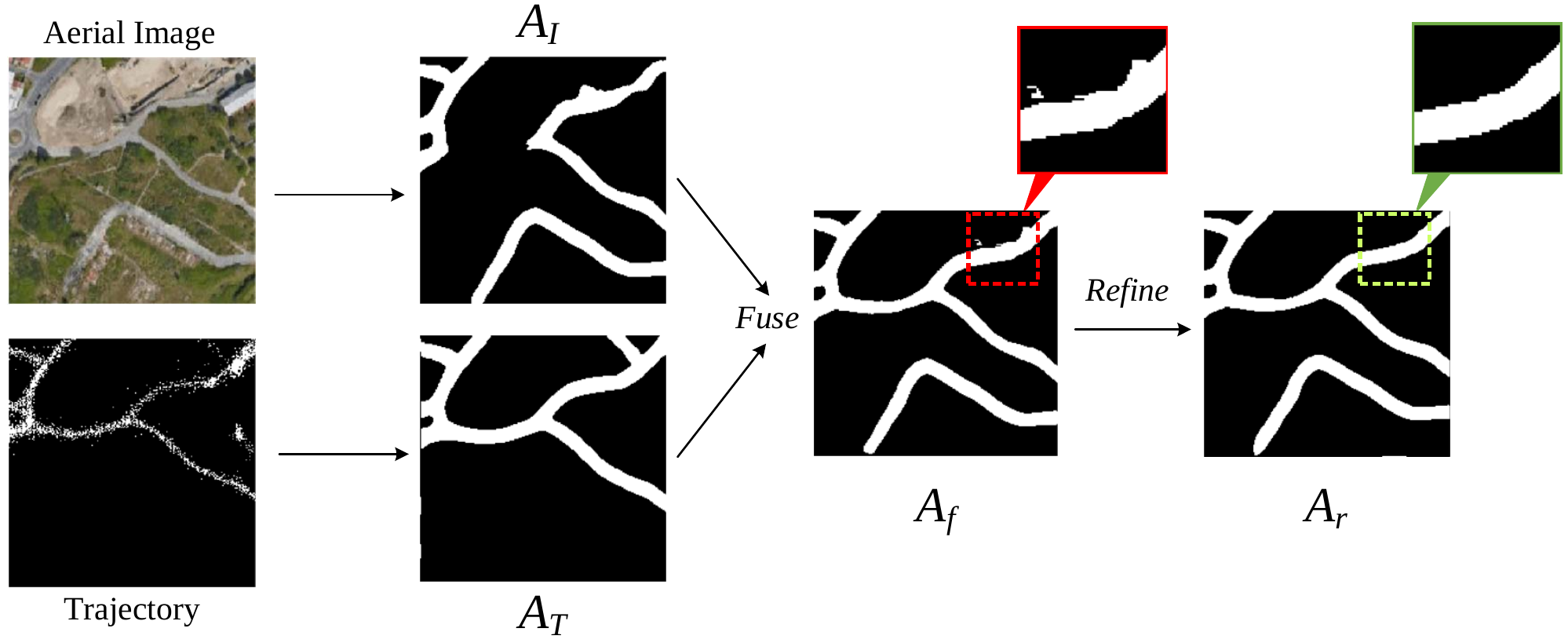}
	\caption{An example showing the process of refinement.}
	\label{fig:refine_sample}
\end{figure}

\paragraph{Fusion Refinement.} 
Recall that GFM outputs the fused features $A_f^{(i)}$ that contain the more useful information from two data sources which are expected to be more confident in making the prediction.
%
%
Inspired by the residual refinement learning \cite{labelrefinement}, here, we directly leverage such information as the base of the feature map and we want the decoder to learn a residual refinement of $A_f^{(i)}$ which can be formalized as the following equation.
\[A_r^{(i)} = A_f^{(i)} + R\left(A_f^{(i)}, A_r^{(i-1)}\right)\]
$A_r^{(i)}$ denotes the refined features of level $i$, based on $A_f^{(i)}$. Here, $R\left(A_f^{(i)}, A_r^{(i-1)}\right)$ is the residual refinement function.
Fig.~\ref{fig:refine_sample} shows an example of the generation of a prediction. First, both image and trajectory modalities make their own independent predictions (with information contained by $A_I^{(i)}$ and $A_T^{(i)}$ respectively), while these predictions might not be sufficient to provide a precise prediction. Next, the GFM fuses these two data sources by enlarging the one being more confident to predict the answer which leads to a more precise prediction. However, we can observe from Fig.~\ref{fig:refine_sample} that such a linear combination may still face some issues, \eg, isolate tiny roads and the un-smoothed predictions. Then, the decoder utilizes the fused feature $A_f^{(i)}$ as the base and learns some area-invariant refinements, such as smoothing the prediction and removing the isolate points/short branches. Such a fusion-refinement process is consistent with our intuition. 

\begin{figure}[t]
	\centering
	\includegraphics[width=70mm]{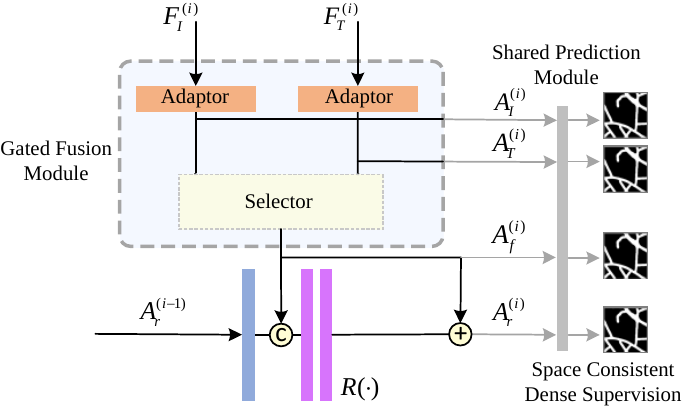}
	\caption{Structure of the densely supervised refinement}
	\label{fig:refinement}
\end{figure}

The refinement decoder still follows the elegant structure of U-Net, 
\ie, 4 decoding blocks composed by a stride 2 deconvolutional layer $D_{2 \times}(\cdot)$ for learnable $2\times$ up-sampling and two $3\times 3$ convolutions (with batch normalization and followed by an ReLu activation), denoted as $CBR^2_{3\times3}(\cdot)$. Concatenation is adopted to combine the information up-sampled from the previous level ($D_{2\times}\left(A_r^{(i-1)}\right)$) and the information from the current level ($A_f^{(i)}$). The difference is that U-Net concatenates the activation maps from encoders by skip-connection while the newly proposed DSRD concatenates the fused features $A_f^{(i)}$ produced by GFM. The following equation formalizes the residual refinement function. Fig.~\ref{fig:refinement} shows the computation flow of the decoder at $i$-th level.
\[R\left(A_f^{(i)}, A_r^{(i-1)}\right) = CBR^2_{3 \times 3}\left(D_{2 \times}\left(A_r^{(i-1)}\right) \textcircled{c} A_f^{(i)}\right)\]




\paragraph{Dense Supervision.} 
We want to highlight that, although we design a refinement structure which seems to be reasonable,
we could not rely on minimizing the final prediction loss as the only objective. If we do so, the refinement structure might not be able to achieve the performance it is designed to achieve, 
%
%
as there is no guidance to ensure that the decoder has indeed learned how to fuse both modalities and how to refine the prediction. Thus, to enforce our refinement procedure to work as expected, we propose a shared prediction module, which can incorporate dense supervisions to explicitly regularize the learned features and to guide the learning procedure.  

Recall that we have introduced two adapters $A_I(\cdot)$ and $A_T(\cdot)$ in GFM to transfer the feature maps $F_I^{(i)}$ and $F_T^{(i)}$ from spaces $\mathcal{F}_I$ and $\mathcal{F}_T$ to a uniform space $\mathcal{A}$. Consequently, the fused feature $A_f^{(i)} = G_I^{(i)} \odot A_I^{(i)} + G_T^{(i)} \odot A_T^{(i)}$, which is the linear combination of $A_I^{(i)}$ and $A_T^{(i)}$, will also lie in $\mathcal{A}$. Besides, the refined feature $A_r^{(i)}$ which is based on $A_f^{(i)}$ (with a learned residual) should also be located in space $\mathcal{A}$.

Accordingly, we decide to adopt a shared prediction module that takes the features in space $\mathcal{A}$ as input and produces a binary prediction. Then, we simultaneously predict the labels from $A_I^{(i)}$, $A_T^{(i)}$, $A_f^{(i)}$ and $A_r^{(i)}$, which involves $5 \times 4 = 20$ supervisions in total. 
The prediction module takes a tensor $A \in \mathcal{A}$ having shape $h^{(i)} \times w^{(i)} \times c^{(i)}$ as the input and generates a binary prediction with the same spatial size, \ie, $h^{(i)} \times w^{(i)}$. It is implemented by an $1\times 1$ convolutional layer $\zeta(\cdot)$ to perform the linear transformation and a 2-classes softmax layer to predict the probability. 

The dense supervision has two functions. 
First, it explicitly constraints space consistence. Recall that we have only \emph{assumed} that the features produced by the adapters lie in the same space without any constraints in the previous sections. However, after adopting the shared prediction module that only accepts one feature space to generate the prediction, the learned space after the adaption and the space of $A_f^{(i)}$ and $A_r^{(i)}$ are forced to be transferred to the uniform space $\mathcal{A}$.
Second, it optimizes the predictions generated by the fused feature $A_f^{(i)} = G_I^{(i)} \odot A_I^{(i)} + G_T^{(i)} \odot A_T^{(i)}$, which is equivalent to optimize the gate values $G_I^{(i)}$ and $G_T^{(i)}$. This supervision actually allows the gradient to be propagated by a short cut to the GFM to facilitate the learning of the fusion process.

\paragraph{Training.}
We adopt pixel-wise cross entropy loss on all the predictions in all levels, i.e., predictions computed from $A^{(i)}_I$, $A^{(i)}_T$, $A^{(i)}_f$ and $A^{(i)}_r$, where $i = 1,2,...,5$. We train the model via Adam optimization \cite{adam} with the learning rate at 1e-4. As our model involves many supervisions, the weights of supervisions $A^{(i)}_I$, $A^{(i)}_T$, $A^{(i)}_f$ are all set to $0.5$ and that of $A^{(i)}_r$ is set to $1.0$ (in all levels). Note that the performance is not very sensitive to the weights. At inference time, $A^{(5)}_r$ is used as the model output.  Please refer to the supplementary material for more training details. 

\section{Experiments}

\subsection{Experiment Setting}

\begin{table*}
	\scriptsize
	\caption{\label{tab:overall_eval} The performance of all approaches under three image-trajectory-paired datasets. TC1, KDE and COBWEB are not learning-based approaches thus their results are deterministic.}
	\begin{center}
		\begin{tabular}{c|cc||cc||cc}
			\hline
			Dataset & \multicolumn{2}{c||}{Porto} & \multicolumn{2}{c||}{Shanghai} & \multicolumn{2}{c}{Singapore}\\
			\hline
			Metric & IoU & F1-score & IoU & F1-score & IoU & F1-score\\
			\hline
			\multicolumn{7}{c}{\textbf{Trajectory-based Approaches}}\\
			\hline
			TC1 & $0.153 \pm 0.000$ & $0.184 \pm 0.000$ & $0.361 \pm 0.000$ & $0.531 \pm 0.000$ & $0.109 \pm 0.000$ & $0.197 \pm 0.000$\\
			KDE & $0.362 \pm 0.000$ & $0.532 \pm 0.000$ & $0.333 \pm 0.000$ & $0.500 \pm 0.000$ & $0.240 \pm 0.000$ & $0.386 \pm 0.000$\\
			COBWEB & $0.277 \pm 0.000$ & $0.434 \pm 0.000$ & $0.334 \pm 0.000$ & $0.501 \pm 0.000$ & $0.196 \pm 0.000$ & $0.328 \pm 0.000$\\
			U-Net$_{traj}$ & $0.643 \pm 0.010$ & $0.783 \pm 0.007$ & $0.498 \pm 0.117$ & $0.657 \pm 0.109$ & $0.552 \pm 0.024$ & $0.711 \pm 0.020$\\
			\hline
			\multicolumn{7}{c}{\textbf{Aerial Image-based Approaches}}\\
			\hline
			DeepRoadMapper & $0.630 \pm 0.007$ & $0.773 \pm 0.005$ & $0.481 \pm 0.017$ & $0.650 \pm 0.016$ & $0.403 \pm 0.015$ & $0.575 \pm 0.015$\\
			RoadTracer & $0.658 \pm 0.012$ & $0.794 \pm 0.009$ & $0.512 \pm 0.023$ & $0.677 \pm 0.020$ & $0.475 \pm 0.013$ & $0.644 \pm 0.012$\\
			SegNet & $0.629 \pm 0.006$ & $0.772 \pm 0.005$ & $0.484 \pm 0.029$ & $0.652 \pm 0.026$ & $0.408 \pm 0.018$ & $0.578 \pm 0.018$\\
			DeconvNet & $0.589 \pm 0.010$ & $0.741 \pm 0.008$ & $ 0.450 \pm 0.011$ & $ 0.617 \pm 0.010$ & $0.265 \pm 0.022$ & $0.418 \pm 0.028$\\
			U-Net & $0.642 \pm 0.013$ & $0.782 \pm 0.010$ &  $0.479 \pm 0.024$ & $0.647 \pm 0.022$ & $0.409 \pm 0.019$ & $0.580 \pm 0.019$\\
			\hline
			\multicolumn{7}{c}{\textbf{Fusion-based Approaches}}\\
			\hline
			Early Fusion & $0.664 \pm 0.020$ & $0.798 \pm 0.015$ & $0.573 \pm 0.027$ & $0.728 \pm 0.021$ & $0.566 \pm 0.031$ & $0.723 \pm 0.025$\\
			Late Fusion & $0.674 \pm 0.005$ & $0.806 \pm 0.003$ & $0.553 \pm 0.027$ & $0.712 \pm 0.022$ & $0.561 \pm 0.027$ & $0.718 \pm 0.022$\\
			SegNet-RC (late fusion + correction) & $0.688 \pm 0.007$ & $0.815 \pm 0.005$ & $0.603 \pm 0.034$ & $0.752 \pm 0.027$ & $0.565 \pm 0.042$ & $0.721 \pm 0.035$\\
			FuseNet (encoder fusion) & $0.667 \pm 0.016$ & $0.800 \pm 0.011$ & $0.558 \pm 0.072$ & $0.714 \pm 0.062$ & $0.599 \pm 0.024$ & $0.749 \pm 0.019$\\
			V-FuseNet (encoder fusion) & $0.688 \pm 0.021$ & $0.815 \pm 0.015$ & $0.626 \pm 0.019$ & $0.770 \pm 0.015$ & $0.605 \pm 0.029$ & $0.754 \pm 0.023$\\
			L3Fsn & $0.675 \pm 0.003$ & $0.806 \pm 0.002$ & $0.603 \pm 0.012$ & $0.752 \pm 0.009$ & $0.555 \pm 0.032$ & $0.713 \pm 0.027$\\
			1D decoder & $0.664 \pm 0.037$ & $0.797 \pm 0.034$ & $0.610 \pm 0.018$ & $0.759 \pm 0.008$ & $0.599 \pm 0.016$ & $0.748 \pm 0.017$ \\
			TriSeg & $0.682 \pm 0.005$ & $0.811 \pm 0.004$ & $0.621 \pm 0.005$ & $0.767 \pm 0.004$ & $0.589 \pm 0.009$ & $0.742 \pm 0.007$\\
			\model & $\mathbf{0.717 \pm 0.003}$ & $\mathbf{0.835 \pm 0.002}$ & $\mathbf{0.634 \pm 0.012}$ & $\mathbf{0.776 \pm 0.009}$ & $\mathbf{0.620 \pm 0.025}$ & $\mathbf{0.765 \pm 0.019}$\\
			\hline
		\end{tabular}
	\end{center}    
\end{table*}

We use three real city-scale datasets containing aerial images and GPS trajectories for evaluation (Porto, Shanghai, and Singapore). 
Following (\citenp{1ddecoder}; \citenp{stackedunet}; \citenp{dlinknet}; \citenp{triseg}), we adopt intersection of union (IoU) and F1-score of road pixels as the evaluation metrics. 
We denote correctly predicted road pixels, \ie, true positive, as TP, correctly predicted non-road pixels, \ie, true negative, as TN, road pixels which are wrongly predicted as non-roads, \ie, false negative, as FN, and non-road pixels which are wrongly predicted as roads, \ie, false positive, as FP. The IoU is computed as $\frac{TP}{TP+FP+FN}$. Precision is computed as $\frac{TP}{TP+FP}$ and recall is computed as $\frac{TP}{TP+FN}$. F1-score, the harmonic mean of precision and recall, is computed as $\frac{2\times precision \times recall}{precision + recall}$. 
The details of the datasets and the metrics are presented in the  supplementary material.

\begin{figure*}[t]
	\centering
	\includegraphics[width=170mm]{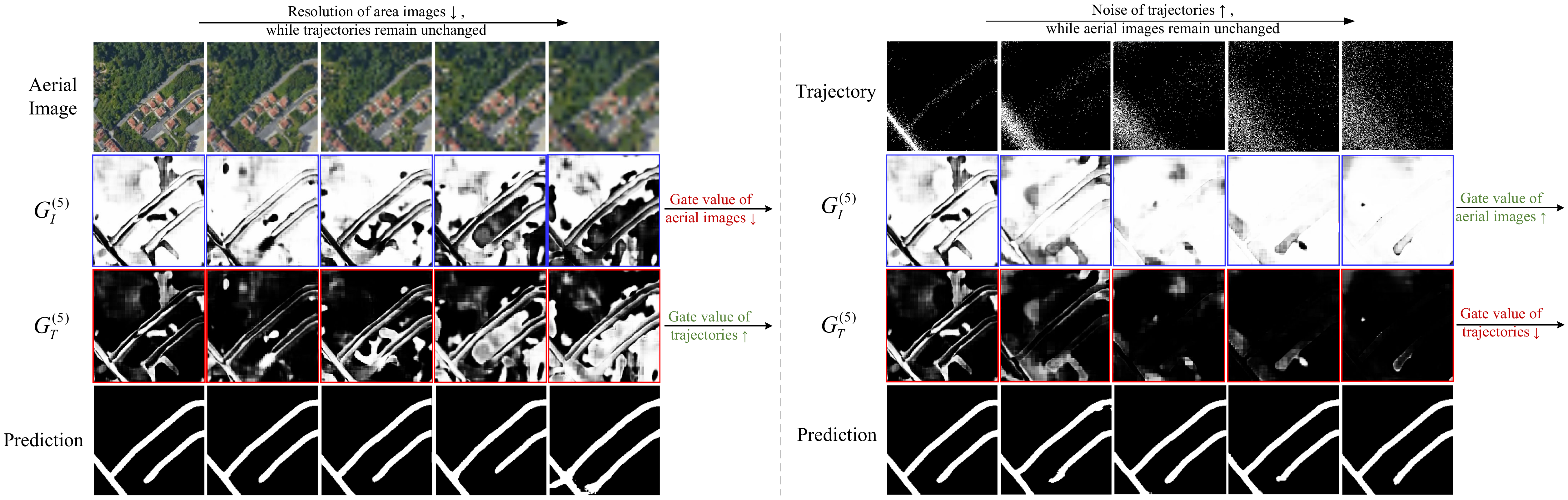}
	\caption{The visualization shows how gate value changes \wrt~one data source being less confident in prediction. The left sample refers to the case where the resolution of an image is reduced gradually, and the right one refers to the case where the localization noises of trajectories are increased. 
		White pixel indicates a large gate value and black pixel indicates a small  value.}\label{fig:gate_dataquality}
\end{figure*}

\begin{figure*}[htb]
	\centering
	\subfigure[Predictions of all data fusion models]
	{\label{fig:atk_result}\includegraphics[width=90mm]{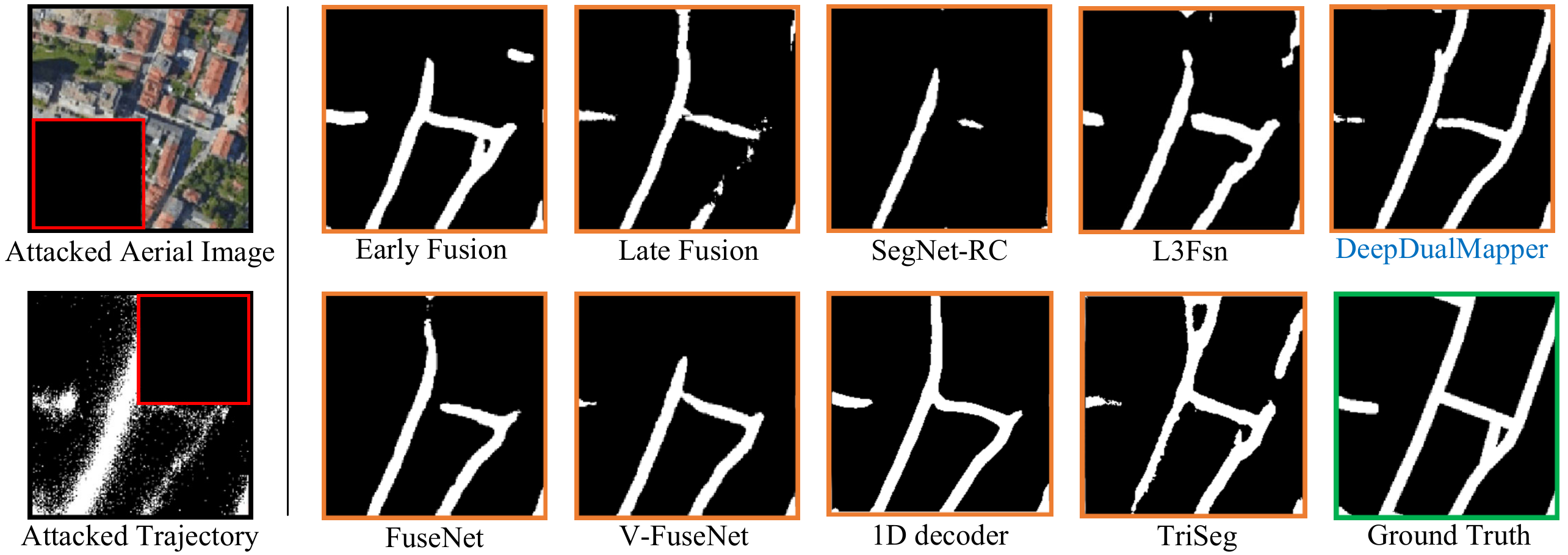}}
	\hspace{0.2in}
	\subfigure[The gate value computed by GFM in all levels]
	{\label{fig:atk_gate}\includegraphics[width=71mm]{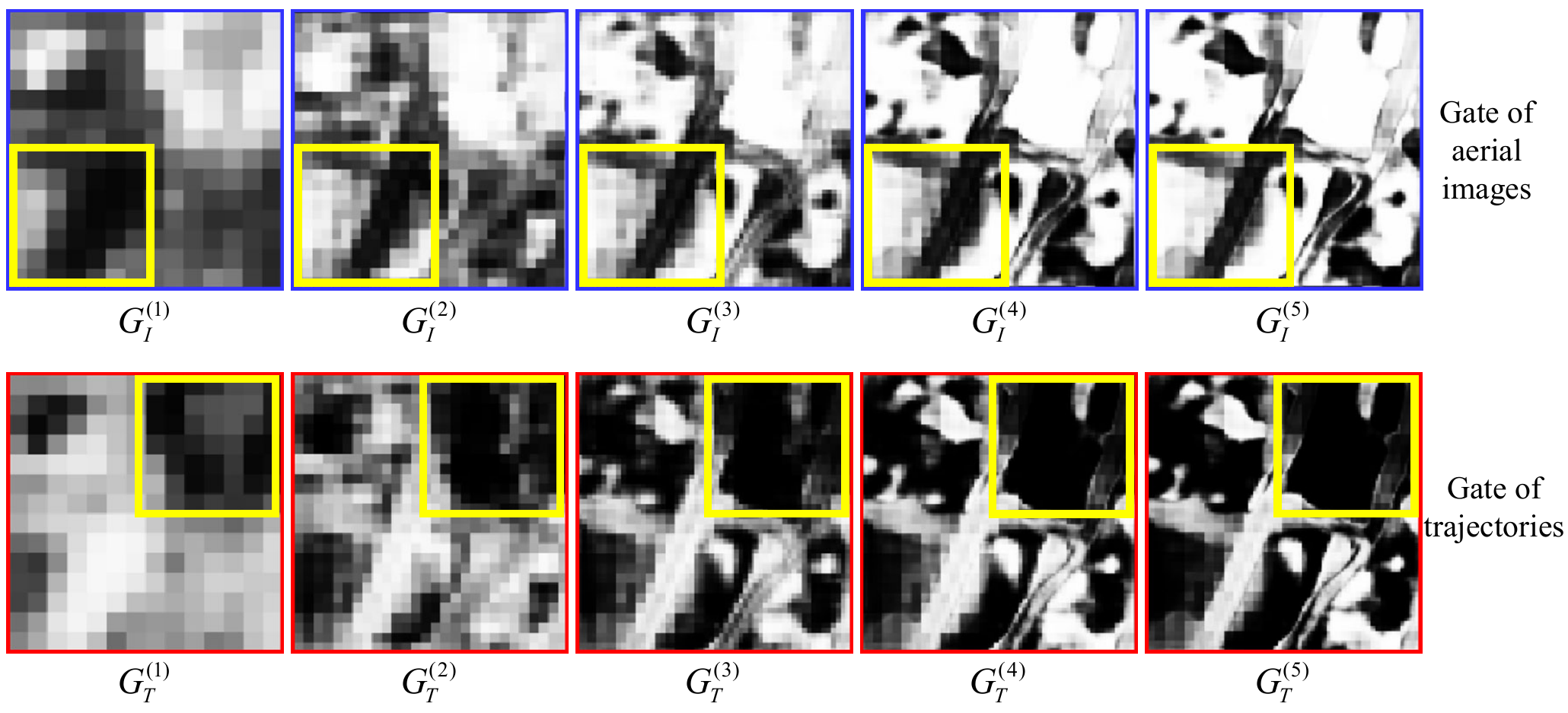}}
	\caption{Visualization of results and gate values under information loss attack. White pixel indicates higher gate value. Pay attention to the areas (in yellow boxes) of gate values which represent the areas attacked by the information loss.}
	\label{fig:atk_vis}
\end{figure*}

\subsubsubsection{Baselines} We incorporate a large number of competitors which can be clustered into three categories. 
The first category is trajectory-based approaches that extract the map based on clustering trajectories/GPS points. Representative works \emph{TC1} \cite{tc1}, \emph{KDE} \cite{kde} and \emph{COBWEB} \cite{cobweb} are selected as the baselines. 
The second category is aerial image-based approaches which extract the map using computer vision techniques. \emph{DeconvNet} \cite{deconvnet} (adopted in \cite{casnet}), \emph{SegNet} \cite{segnet} (adopted in \cite{robocodes}), \emph{U-Net} \cite{unet},
\emph{DeepRoadMapper} \cite{deeproadmapper} and \emph{RoadTracer} \cite{roadtracer} are implemented as the competitors. 
%
We also include the U-Net which takes in trajectory data, pre-processed as an image, as input, denoted as \emph{U-Net$_{traj}$}. 
The last category is fusion-based approaches which take both aerial image and trajectory as inputs. We implement \emph{Early-fusion} which concatenates the inputs (using U-Net), \emph{L3Fsn} concatenating the features in the third convolutional block of FCN-8 \cite{l3fsn}, \emph{Late-fusion} and the version with residual correction namely \emph{SegNet-RC}\cite{segnetrc}, and \emph{FuseNet} \cite{fusenet} as well as its advanced version \emph{V-FuseNet} applied in (\citenp{beyondrgb}; \citenp{anotherfusenet}; \citenp{rgbosmfuse}). \emph{TriSeg} \cite{triseg} and \emph{1D Decoder} \cite{1ddecoder}, the only two existing state-of-the-art fusion approaches for map extraction task, are also implemented as the competitors.

\subsection{Overall Evaluation}
We evaluate the performance of different approaches in extracting the maps from three cities under test regions. The results are reported in Table~\ref{tab:overall_eval}.
First, most trajectory-based approaches do not perform well. This is consistent with our expectation, as they are based on trajectory clustering without any supervised learning procedure. U-Net$_{traj}$ performs best among them as it is trained end-to-end. 
For those image-based approaches, DeepRoadMapper and RoadTracer perform better than the others in general as they use specifically designed CNN structure which is more powerful than other VGG-like models.
In general, fusion-based approaches outperform the approaches in the above two categories. It demonstrates the power of combining different data sources and justifies that the combination of data sources that complement each other can significantly improve the overall performance. 
\model demonstrates its superior performance, as it consistently outperforms all the competitors in all three cases. Note that \model achieves a much stabler performance as its standard deviation tends to be small. Note that to prove the effectiveness of our fusion strategy and to assure the fairness of our evaluation, all the approaches evaluated use the same input features. 
Among other fusion-based approaches, TriSeg, 1D decoder, and V-FuseNet demonstrate certain advantages over others; while the performance ranking of remaining approaches is not clear.
%

Notice that among these three datasets, the performance on the Porto dataset is the best. The main reason is that Porto dataset has better data quality than others. Porto's satellite image is more accurate and clearer than the other two. For trajectory data, the noise and error on trajectories of Porto are smaller than that of Singapore, and the coverage density is higher than that of Shanghai. Consequently, the overall performance of Porto is the best of all three datasets.

\subsection{Visualization of the Gate Values}
To offer a clearer view of how our fusion model behaves when it encounters two data sources with different confidence of predictability, we select a sub-region and gradually reduce the predictability in either modality. We report the gate values $G_I^{(5)}$ and $G_T^{(5)}$ of level 5 (\ie~the last level) in Fig.~\ref{fig:gate_dataquality} to visualize the changes to the information flow when one data source's quality drops. We can infer that when the image fades and starts to lose more and more details, it gradually loses its confidence in prediction. Therefore, $G_I^{(5)}$ is reduced (becomes blacker) while $G_T^{(5)}$ is increased (becomes whiter). However, it is still able to make a reasonable prediction even when the original aerial image has lost much important information. 
Similarly, we can make the same observations from the case where the noise of trajectory data increases. 

\subsection{Performance w.r.t. Trajectory Data Volume}
\begin{figure}[t]
	\centering
	\includegraphics[width=50mm]{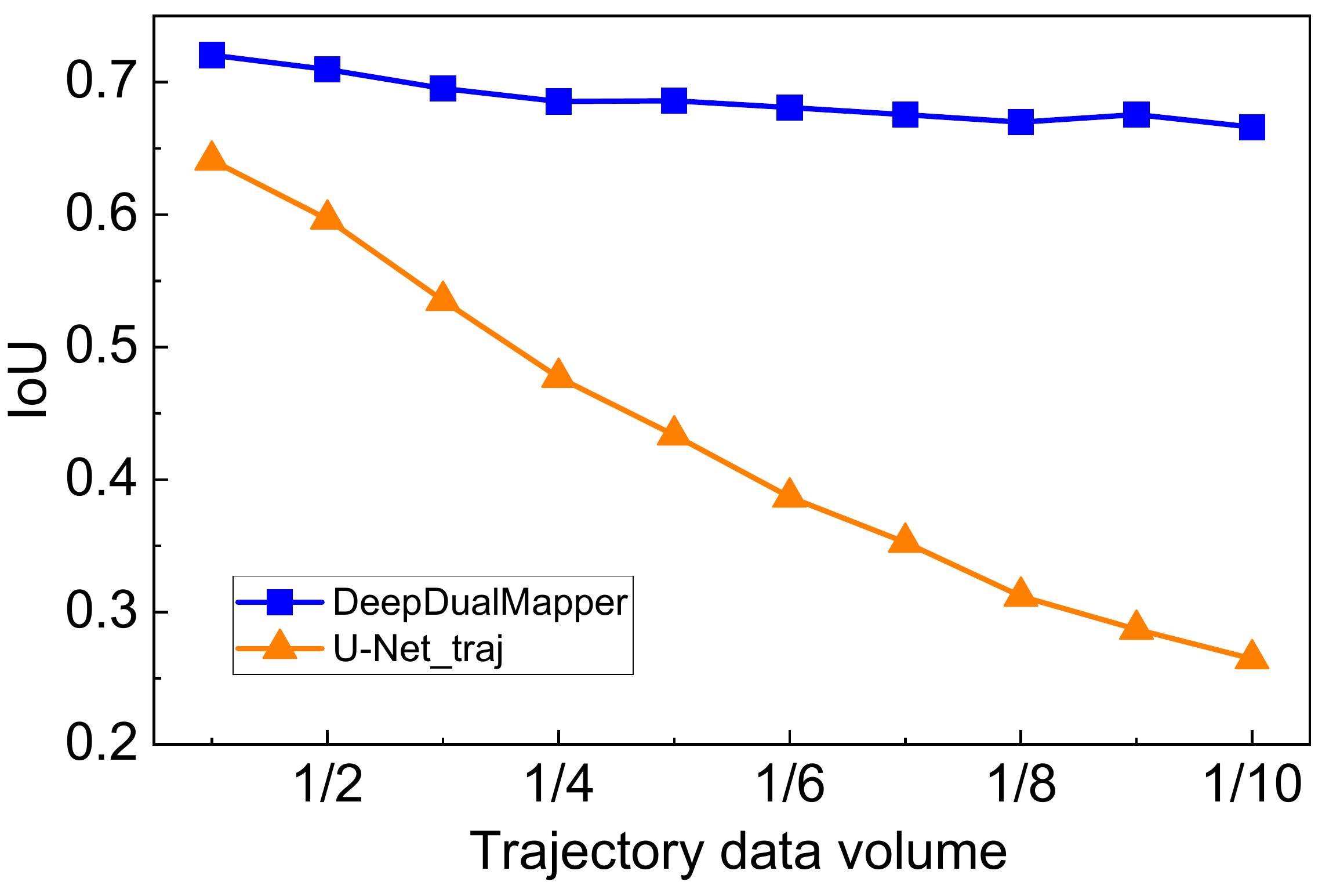}
	\caption{The IoU of DeepDualMapper (dual modality) and U-Net$_{traj}$ (single modality) under Porto dataset w.r.t. different trajectory data qualities.}
	\label{fig:data_quantity}
\end{figure}
Fig.~\ref{fig:data_quantity} plots the IoU performance of U-Net$_{traj}$ and DeepDualMapper under different trajectory data volumes. We only report the results of the Porto dataset for space saving, while similar results are observed from the other two datasets. U-Net$_{traj}$ is selected for comparison because it is the backbone of our model and it can be directly compared with DeepDualMapper that takes in multiple data sources as input. It is observed that U-Net$_{traj}$ is vulnerable to the reduction of trajectory data quality, while DeepDualMapper demonstrates resilience to the reduction of data quality. Although DeepDualMapper is relatively robust to data quality, it still has a minor performance drop. We think it is reasonable. When one modality (i.e.. trajectory data) loses most of its information, DeepDualMapper can only infer the information from the other modality (i.e., aerial image) with its performance similar to the U-Net of single modality (i.e., aerial image).

\subsection{Evaluation on Robustness to Information Loss}
Data fusion approaches will be effective only if they own the capability of learning how to compensate for the information loss of one modality by other modalities. As stated before, existing approaches roughly fuse the data sources so it remains unknown whether they have really learned how to correctly fuse the data and to maximize the utility of both data sources. 
%
%
To have a more thorough study, we conduct the information loss attack on \emph{testing} datasets. In detail, we randomly remove the information of 1/4 area of the aerial image and another 1/4 area of the trajectory data, to see whether the fusion model can correctly find out the right modality containing the true information to perform the inference. We conduct the experiment in all three datasets and report the quantitative results in Table~\ref{tab:dataloss_atk}. In addition, we visualize the maps generated by all data fusion models in Fig.~\ref{fig:atk_result}. Note that quadrant 3 of the aerial image and quadrant 1 of the trajectory data have been removed. 
Most of the baselines are vulnerable to the information loss attack while our model, to some extent, is able to defend such attack.

To further study how \model fuses the data and why it outperforms the other approaches under an information loss attack, we visualize the gate values computed by GFM from level 1 to level 5 in Fig.~\ref{fig:atk_gate}. We can observe that since quadrant 3 of the aerial image has been removed, 
%
GFM assigns higher weights to the trajectory data.
Similarly, as quadrant 1 of the trajectory data has been removed, GFM detects that the aerial image carries more valuable information for the prediction. Accordingly, it passes the information of aerial image but blocks the trajectory information in quadrant 1.
%
%
This example demonstrates the effectiveness of our gating mechanism for data fusion. 
In addition, from the gate values across different levels, we can observe that the gate values are refined from coarse to fine and the gate values of all levels meet the consistency constraint. 

\begin{table}
	\scriptsize
	\caption{\label{tab:dataloss_atk} \textls[-8]{The reported IoU under the test set with information loss attack. Note we do not include the attack in training set.}}
	\begin{center}
		\begin{tabular}{c|ccc}
			\hline
			Dataset & Porto & Shanghai & Singapore\\
			\hline
			Early Fusion & $0.500\pm0.023$ & $0.440\pm0.014$ & $0.420\pm0.016$ \\
			Late Fusion & $0.475\pm0.011$ & $0.407\pm0.059$ & $0.379\pm0.035$ \\
			SegNet-RC & $0.546\pm0.025$ & $0.442\pm0.028$ & $0.393\pm0.026$ \\
			FuseNet & $0.463\pm0.042$ & $0.455\pm0.012$ & $0.405\pm0.039$\\
			V-FuseNet & $0.530\pm0.028$ & $0.432\pm0.043$ & $0.445\pm0.016$ \\
			L3Fsn & $0.510\pm0.027$ & $0.450\pm0.011$ & $0.391\pm0.045$\\
			1D decoder & $0.471\pm0.043$ & $0.455\pm0.005$ & $0.411\pm0.037$\\
			TriSeg & $0.462\pm0.030$ & $0.467\pm0.022$ & $0.398\pm0.024$ \\
			\model & $\mathbf{0.588\pm0.019}$ & $\mathbf{0.497\pm0.023}$ & $\mathbf{0.477\pm0.030}$ \\
			\hline
		\end{tabular}
	\end{center}   
\end{table}

\subsection{Study on Densely Supervised Refinement}
\begin{figure}[t]
	\centering
	\subfigure[IoU predicted by $A_I$, $A_T$, $A_f$ and $A_r$]
	{\label{fig:refinement_eval}\includegraphics[width=40mm]{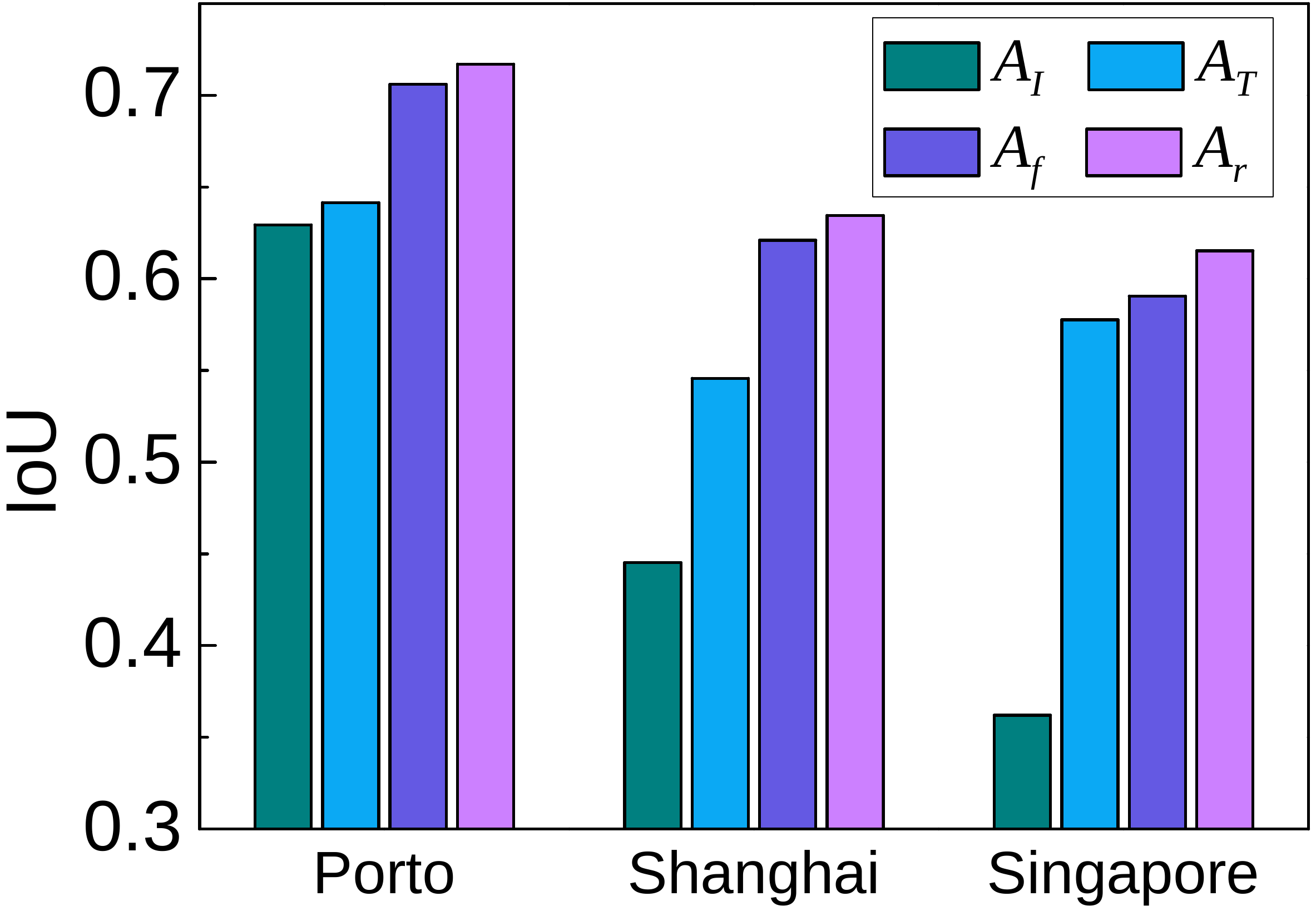}}
	\subfigure[IoU of $A_f$ vs. supervision]
	{\label{fig:supervision_eval}\includegraphics[width=40mm]{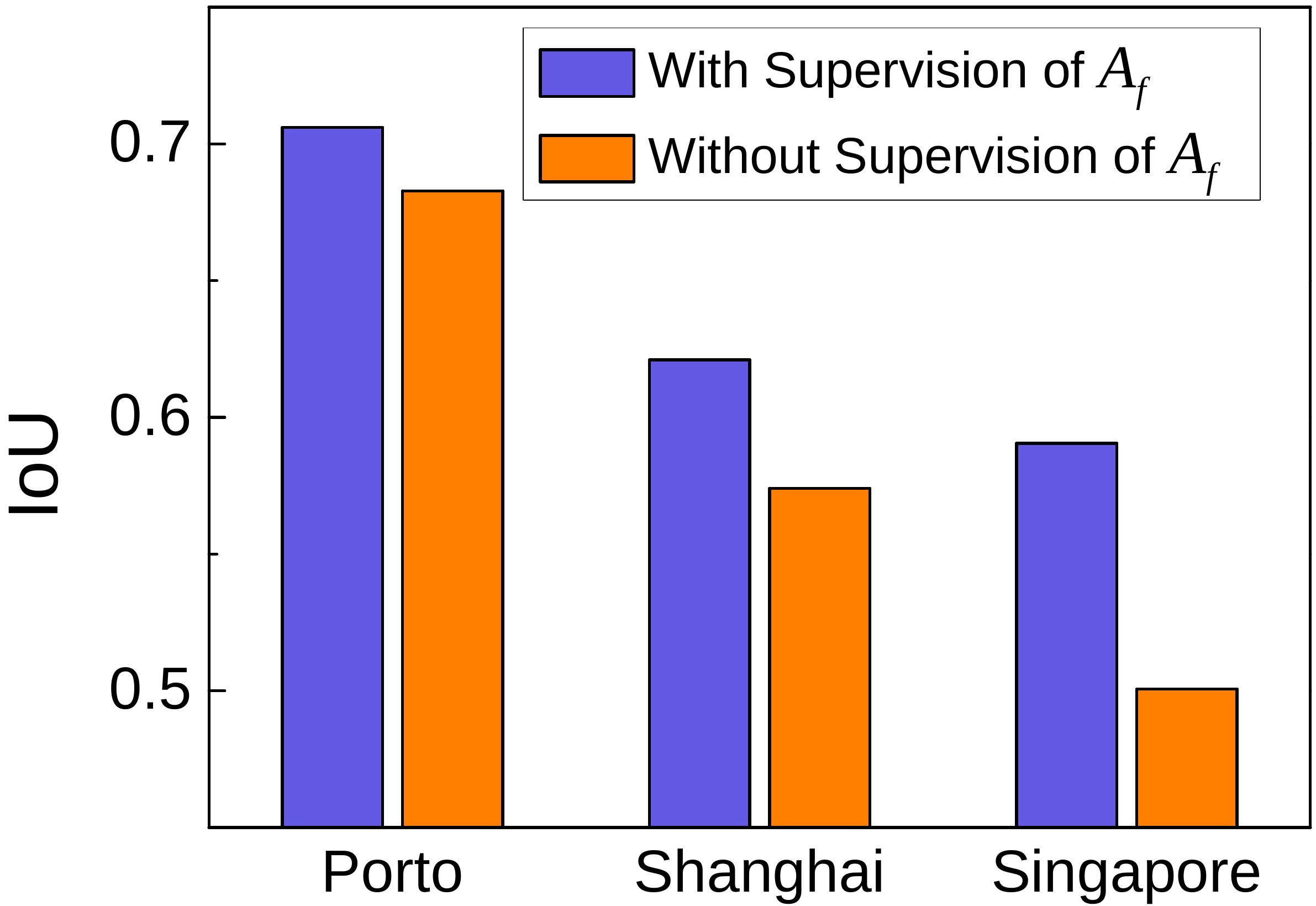}}
	\label{fig:refinement_exp}
	\caption{Quantitative results showing the utility of densely supervised refinement decoding. The IoU of predictions generated by features at each stage are gradually increased demonstrating the effectiveness of the fusion step and the refinement step.}
\end{figure}

\begin{figure}[t]
	\centering
	\includegraphics[width=80mm]{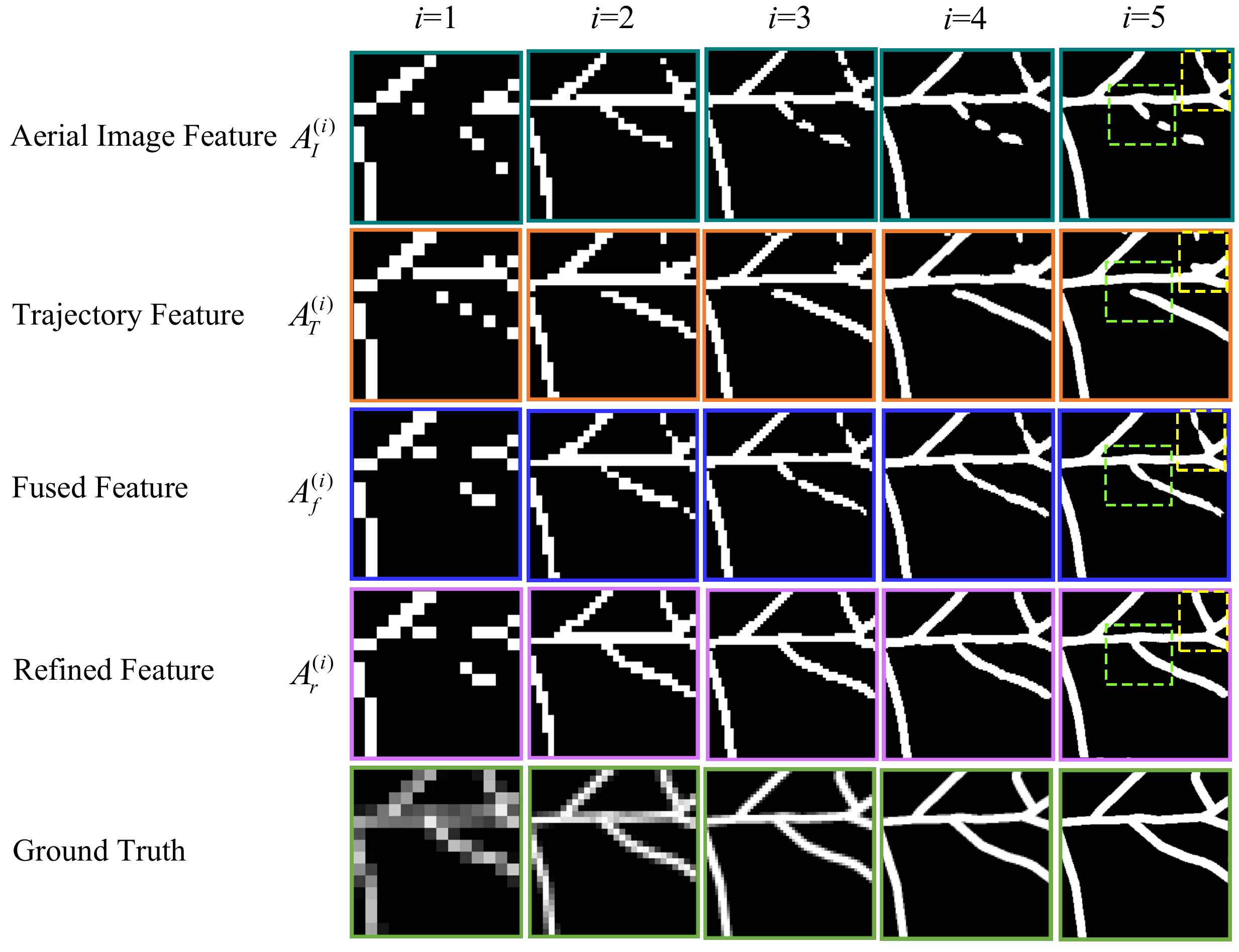}
	\caption{Visualizations of the predictions generated by $A_I$, $A_T$, $A_f$ and $A_r$ in all 5 levels. The fused predictions are better than those of $A_I$ and $A_T$. The refined predictions further smooth (and connect) the roads predicted by $A_f$. Please pay attention to the differences in the dotted boxes.}
	\label{fig:aux_vis}
\end{figure}

Recall that the DSRD is designed to be responsible for generating the predictions. We first evaluate the IoU performance of the predictions generated by the features of $A_I^{(5)}$ (image feature), $A_T^{(5)}$ (trajectory feature), $A_f^{(5)}$ (fused feature) and $A_r^{(5)}$ (refined feature) via the shared prediction module. The results are shown in Fig.~\ref{fig:refinement_eval}. As compared with the information captured by the features $A_I^{(5)}$ or $A_T^{(5)}$ of single modality, the fused feature $A_f^{(5)}$ contains more precise information. In addition, we are able to observe the improvement achieved by $A_r^{(5)}$ over $A_f^{(5)}$, which demonstrates the effectiveness of the refinement process. Last but not least, we claim that supervision can further improve the fusion process as the prediction made by the fused feature $A_f$ with the dense supervision is more accurate than that without the supervision, as reported in Fig.~\ref{fig:supervision_eval}. 
%
%
For a clearer view, we visualize the predictions computed by the features of $A_I^{(i)}$, $A_T^{(i)}$, $A_f^{(i)}$ and $A_r^{(i)}$ in all 5 levels in Fig.~\ref{fig:aux_vis}. From the results, we can observe that the refinement procedure smooths and connects the roads and the result is generated through the feature maps from coarse to fine as well as from rough to smooth over five levels.

	\section{Conclusion}
	In this paper, we have presented an automatic map extraction approach that can effectively fuse the information of aerial images and that of GPS trajectories. We further boost the performance and accuracy of map extraction task by proposing the gated fusion module and the densely supervised refinement decoder.  
	We have demonstrated the effectiveness of our model through comprehensive experiment studies based on three city-scale datasets. In addition, we have implemented an information loss attack task. As expected, our model is much more robust to the attack, compared with a wide range of state-of-the-art competitors, with its resilience to the information loss mainly contributed by its delicately designed fusion structure.
	Since our method is still relatively simple for the preprocessing of trajectory data, we plan to consider other information such as speed and direction in the future and investigate whether performing more data augmentation could enhance the performance of the model.

	\section{Acknowledgements}
	We thank Zhangqing Shan for providing experimental results of trajectory-based approaches (i.e., TC1, KDE, COBWEB). This research is supported in part by the National Natural Science Foundation of China under grant 61772138, the National Key Research and Development Program of China under grant 2018YFB0505000, and the National Research Foundation, Prime Minister's Office, Singapore under its International Research Centres in Singapore Funding Initiative.
	
	\clearpage
	{
		\bibliographystyle{aaai}
		\bibliography{ref}
	}

    \onecolumn
\begin{center}
	\textbf{\LARGE Supplementary Materials for DeepDualMapper}
\end{center}

\vspace{0.2in}

\appendix
\section{Quantitative Results}
\begin{figure}[htb]
	\centering
	\subfigure[Aerial Image]
	{\label{fig:image1}\includegraphics[width=40mm]{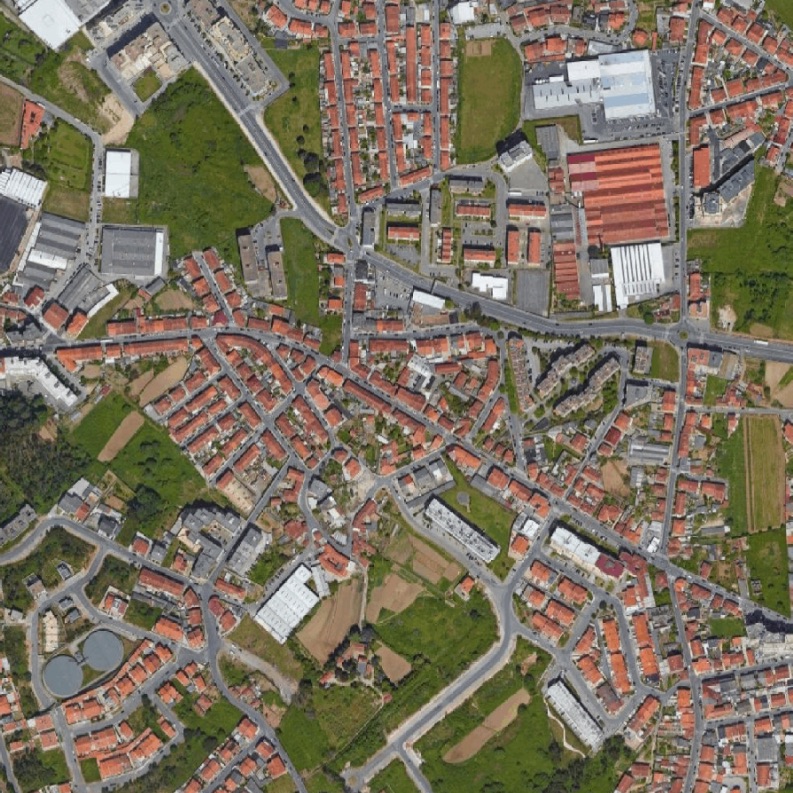}}
	\hspace{0.05in}
	\subfigure[Trajectory]
	{\label{fig:image1}\includegraphics[width=40mm]{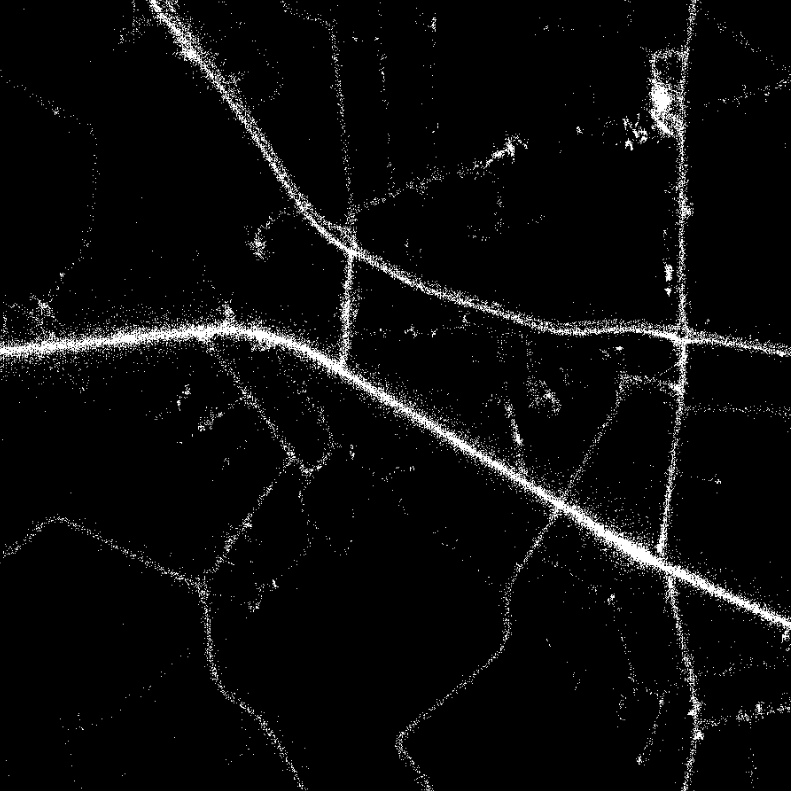}}
	\hspace{0.05in}
	\subfigure[DeepDualMapper]
	{\label{fig:image1}\includegraphics[width=40mm]{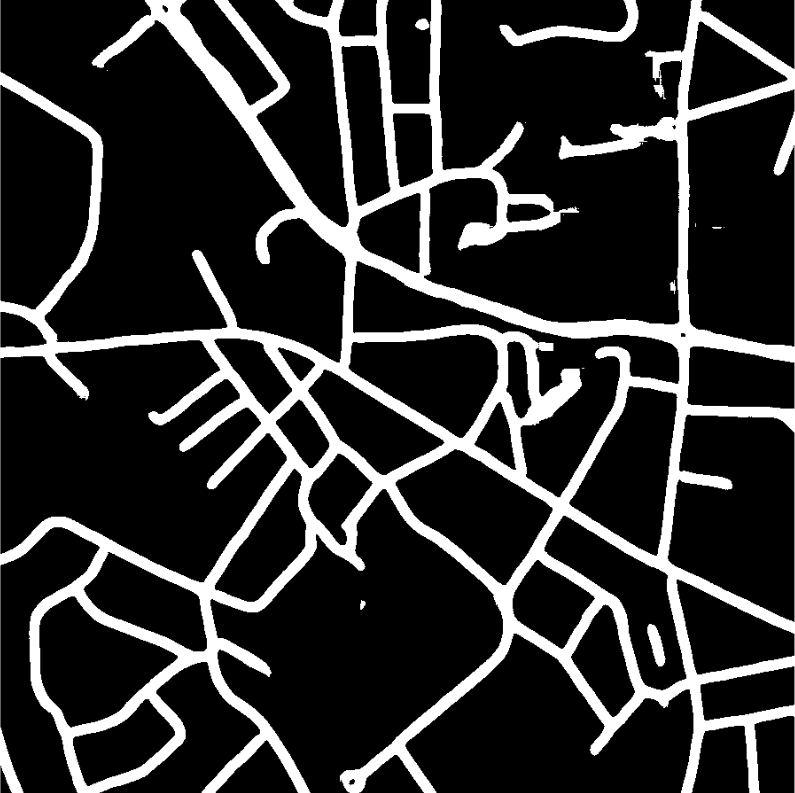}}
	\hspace{0.05in}
	\subfigure[Ground Truth]
	{\label{fig:image1}\includegraphics[width=40mm]{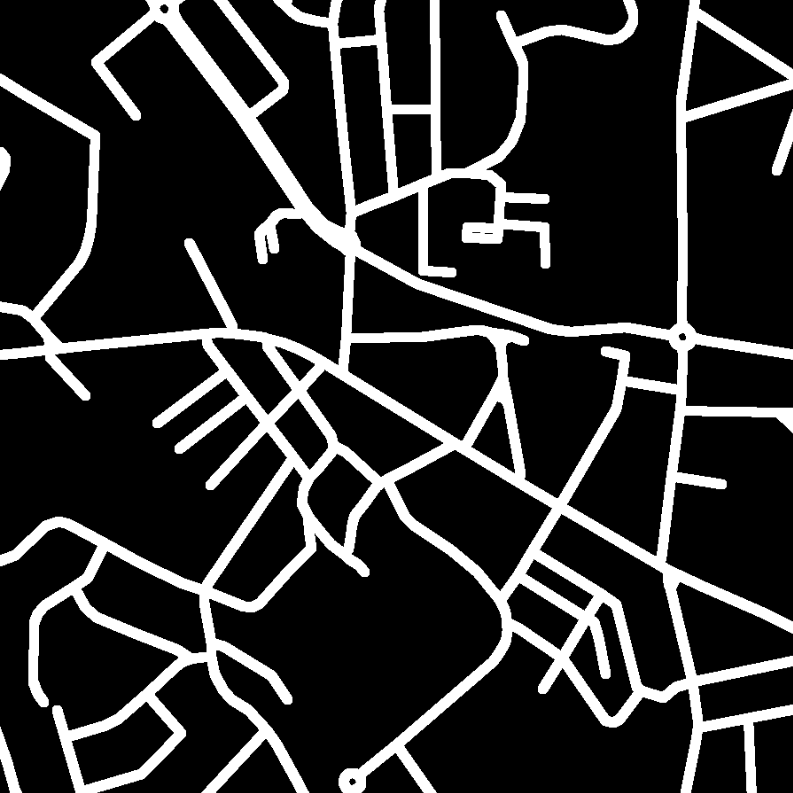}}\\	
	\subfigure[DeepDualMapper(IoU 0.79)]
	{\label{fig:image1}\includegraphics[width=40mm]{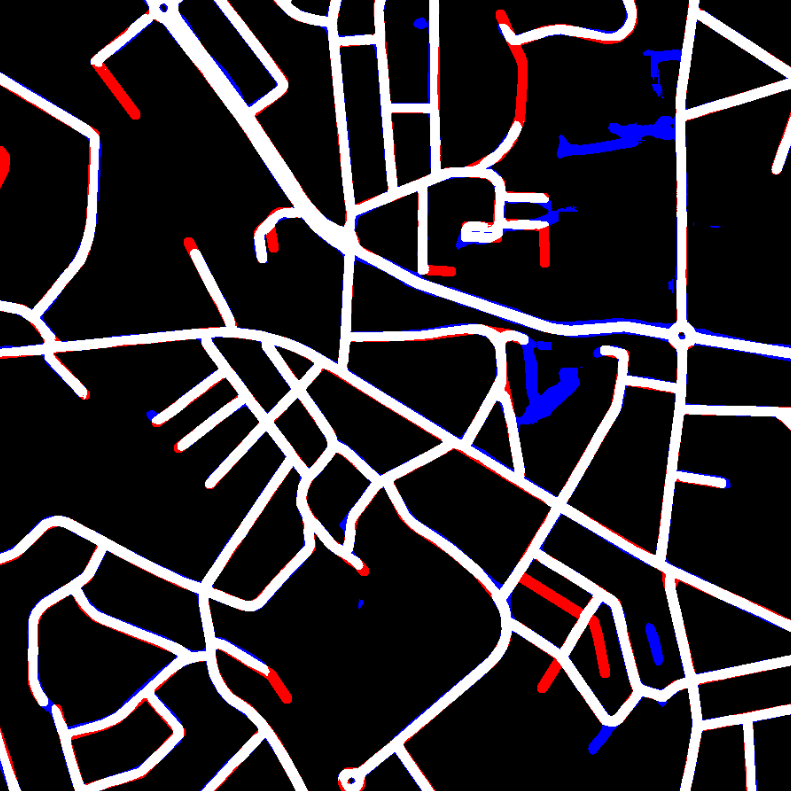}}	
	\hspace{0.05in}
	\subfigure[TriSeg (IoU 0.755)]
	{\label{fig:image1}\includegraphics[width=40mm]{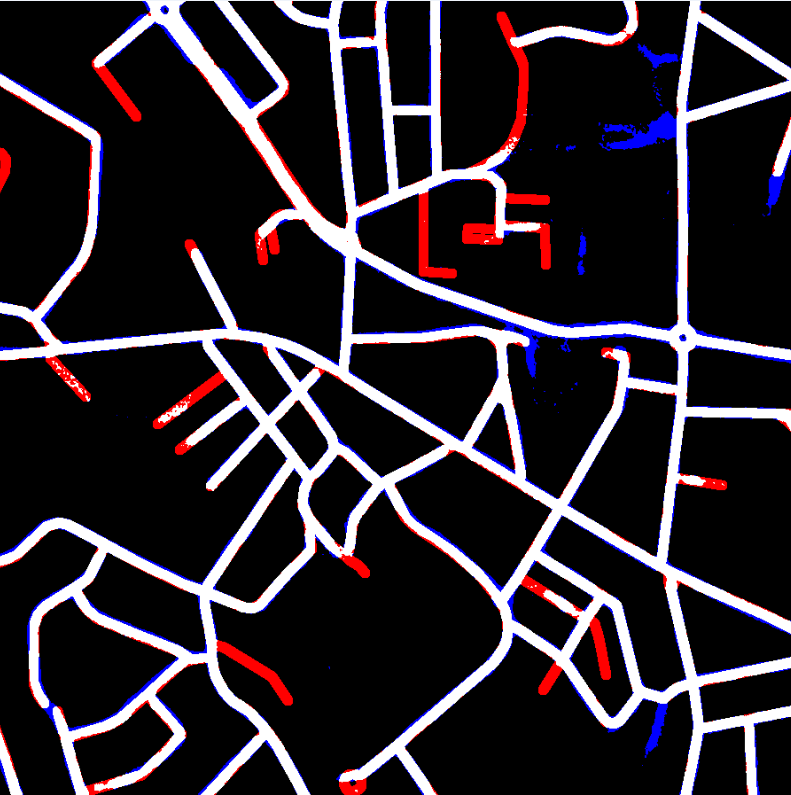}}	
	\hspace{0.05in}
	\subfigure[1D decoder (IoU 0.719)]
	{\label{fig:image1}\includegraphics[width=40mm]{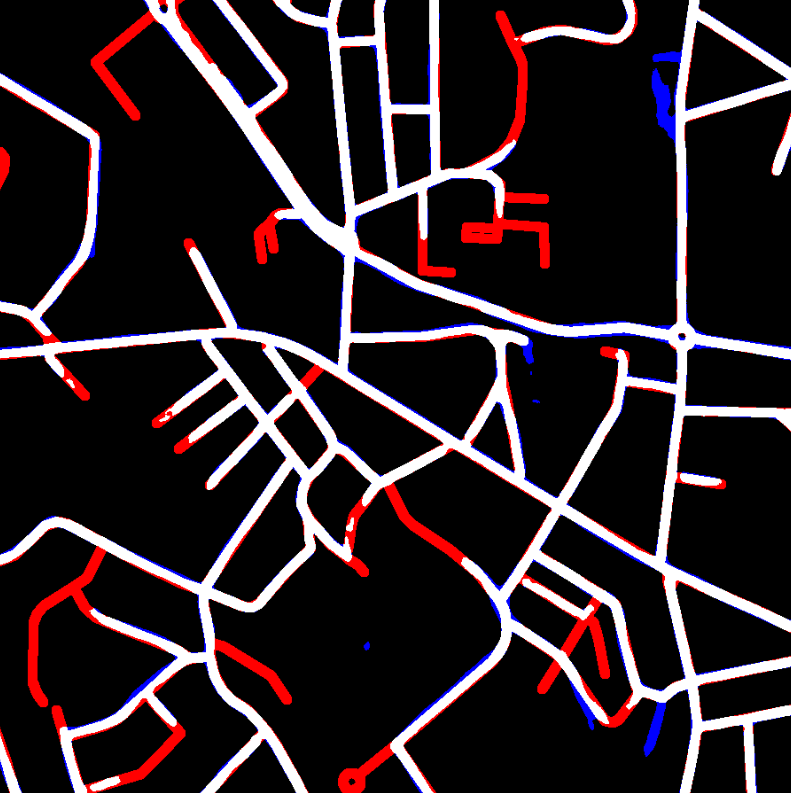}}	
	\hspace{0.05in}
	\subfigure[V-Fusenet (IoU 0.722)]
	{\label{fig:image1}\includegraphics[width=40mm]{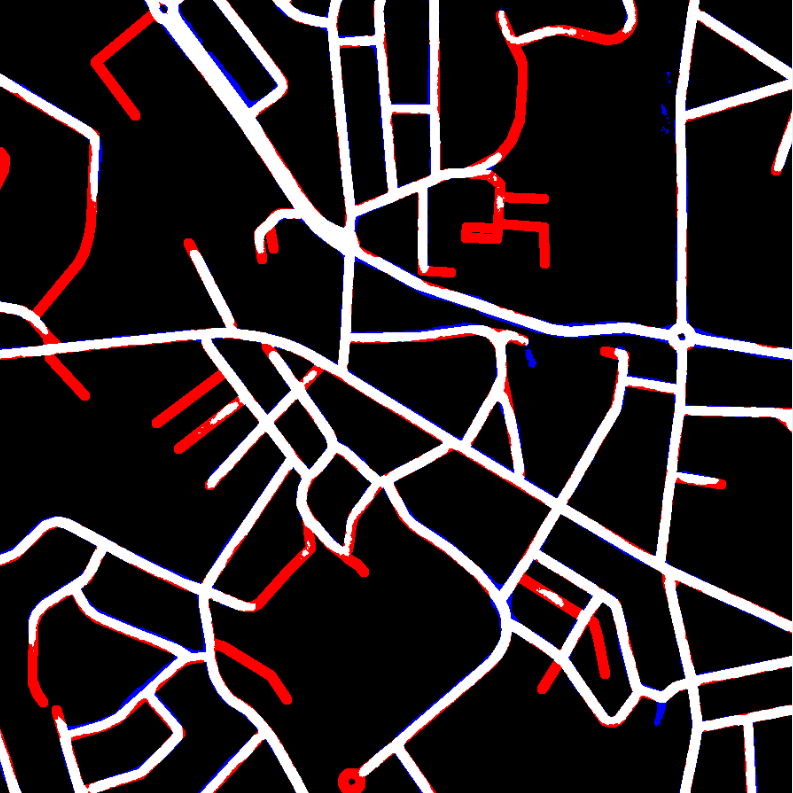}}\\	
	\hspace{0.05in}
	\subfigure[Early-fusion (IoU 0.714)]
	{\label{fig:image1}\includegraphics[width=40mm]{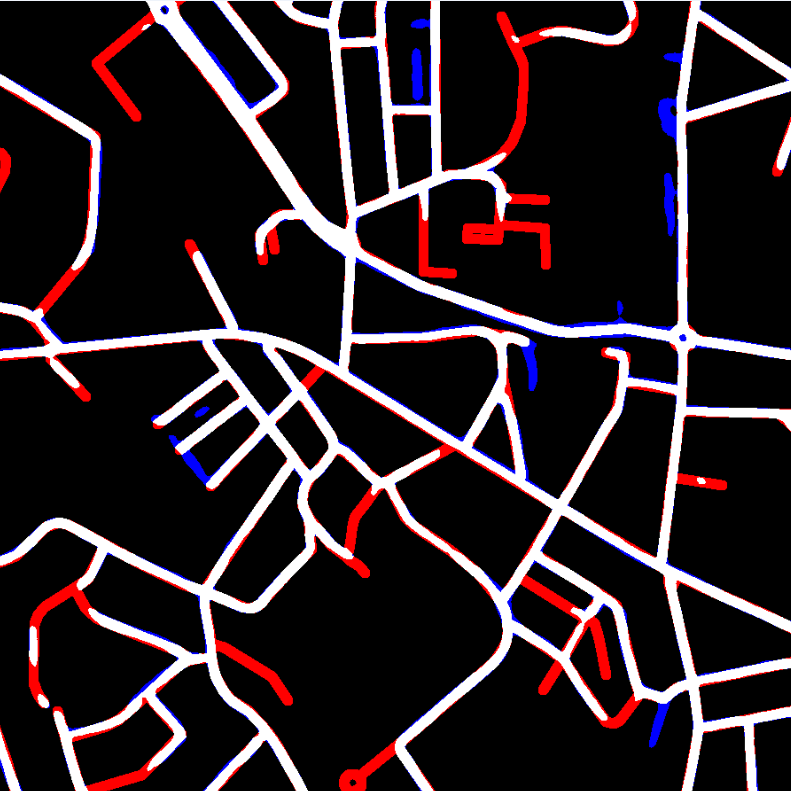}}	
	\hspace{0.05in}
	\subfigure[Late-fusion (IoU 0.747)]
	{\label{fig:image1}\includegraphics[width=40mm]{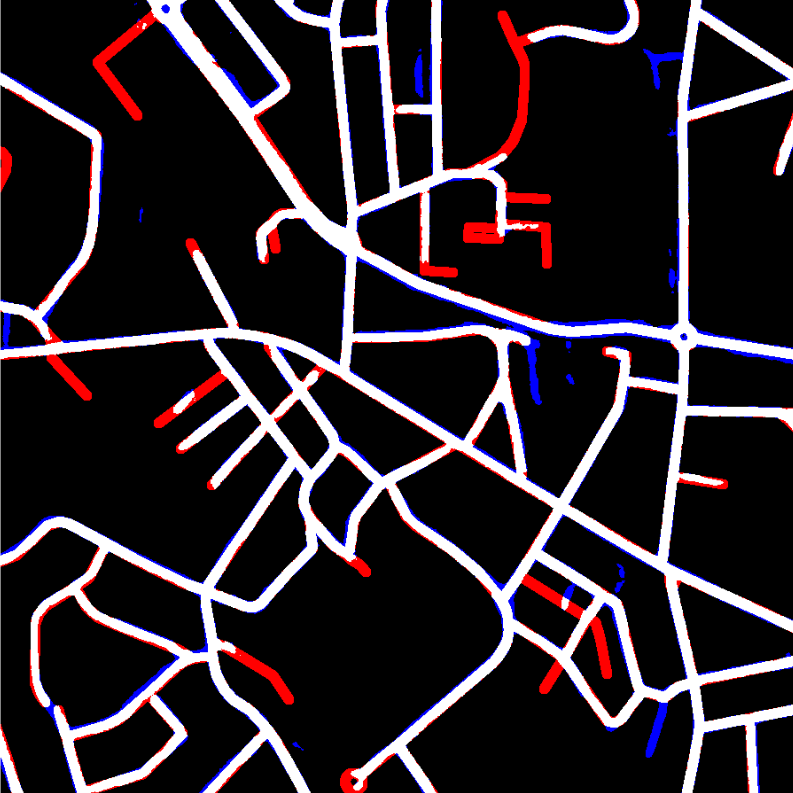}}	
	\hspace{0.05in}
	\subfigure[SegNet-RC (IoU 0.728)]
	{\label{fig:image1}\includegraphics[width=40mm]{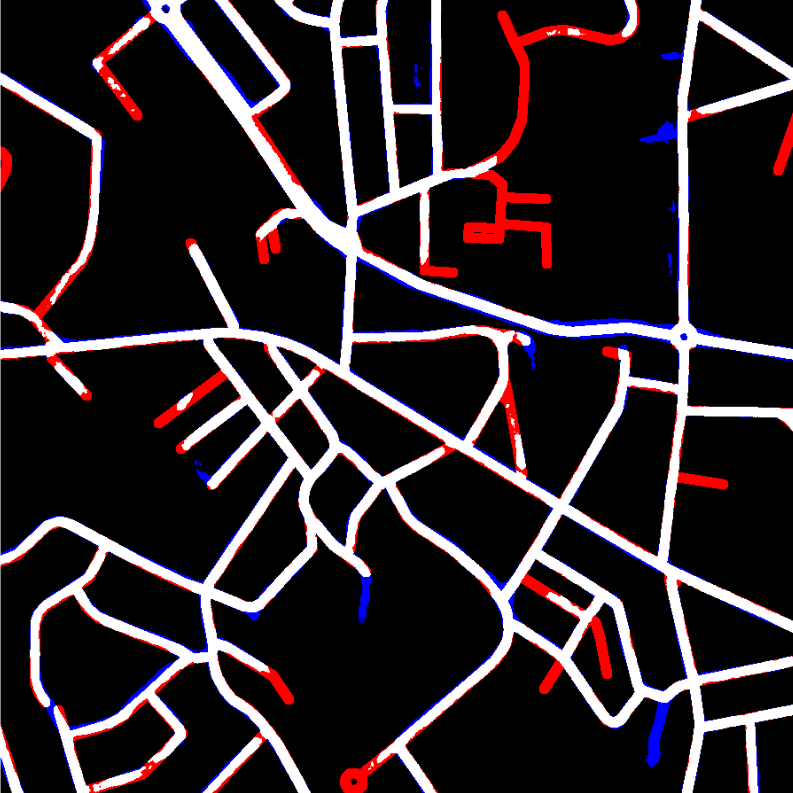}}	
	\hspace{0.05in}
	\subfigure[L3Fsn (IoU 0.708)]
	{\label{fig:image1}\includegraphics[width=40mm]{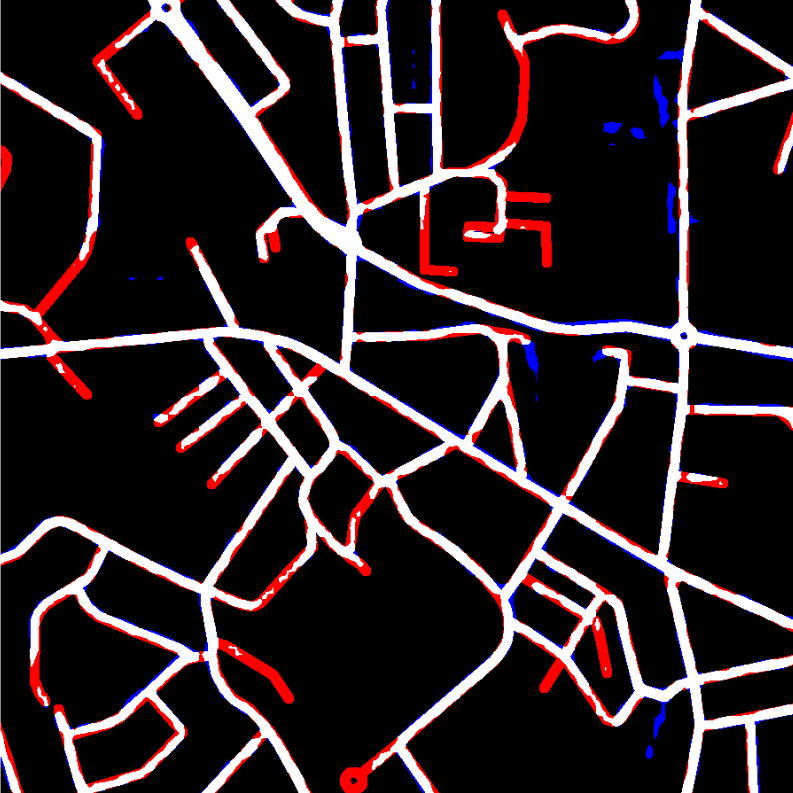}}	
	\caption{Examples showing the generated map in an area of size 900m $\times$ 900m. The first row shows the aerial image and the trajectories as well as the predictions from \model and the ground truth constructed by OpenStreetMap. The second and third rows visualize the predictions of all fusion-based methods. The red pixels indicate the road pixels that are wrongly predicted as non-roads, i.e., false negative, and the blue pixels represent the non-road pixels wrongly predicted as roads corresponding to the ground truth, i.e., false positive.}
	\label{fig:exp_suppl}
\end{figure}

\begin{figure*}[t]
	\centering
	\includegraphics[height=100mm]{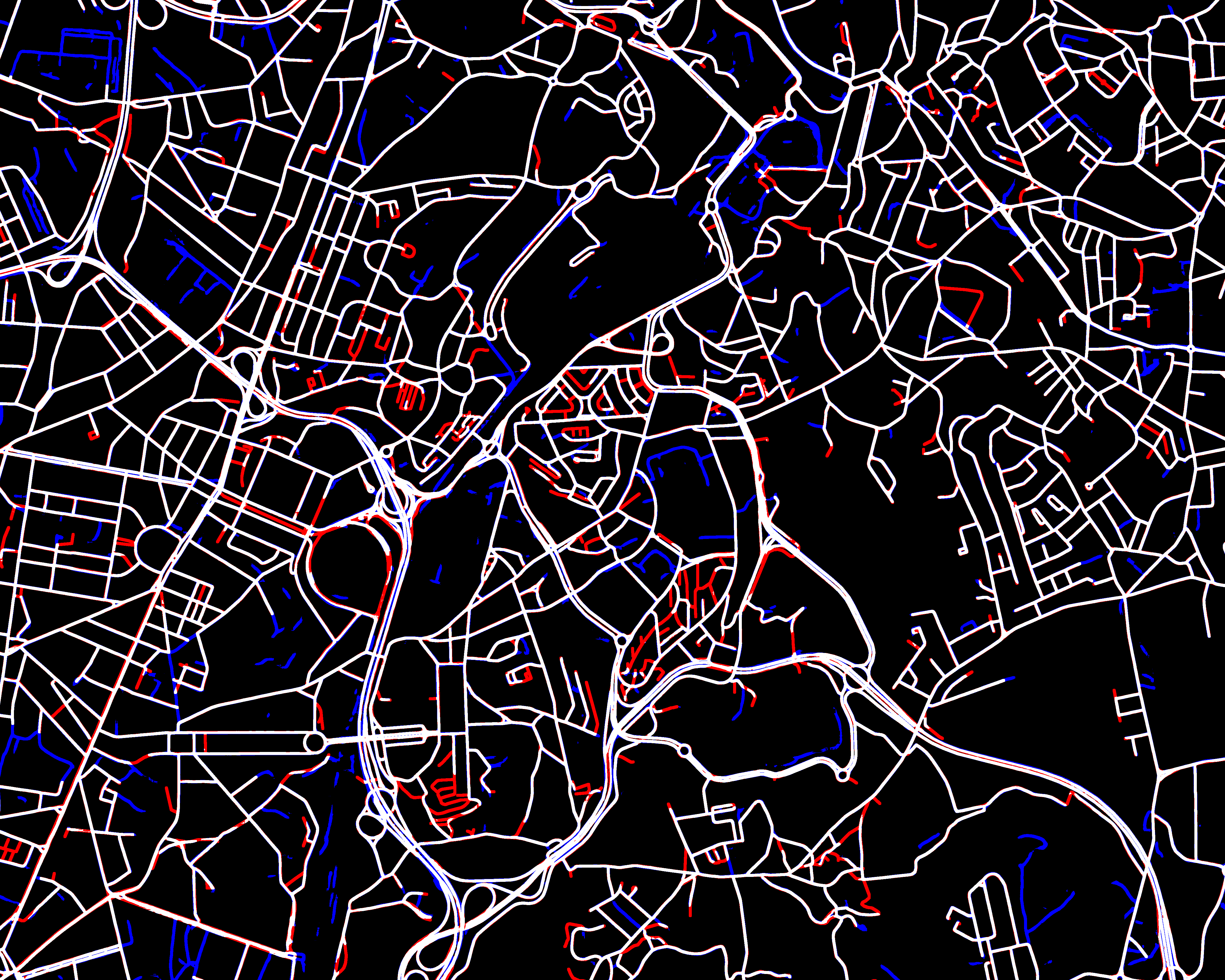}
	\caption{Visualization of the generated map by \model w.r.t. whole test area in Porto dataset. Similarly, the red pixels indicate the pixels being wrongly predicted as non-roads (false negative) and the blue pixels represent the non-road pixels wrongly predicted as roads (false positive). Note that we have found that a large ratio of those false positive samples (in blue) is actually resulted by the incorrectness of OpenStreetMap (served as the ground truth) which means these roads are actually missing in OpenStreetMap. It demonstrates that \model can reversely detect some missing roads in OpenStreetMap and helps to make it more precise.}
	\label{fig:porto_test_vis}
\end{figure*}

We first visualize the generated maps produced by all fusion-based methods introduced in the main text. The generated area has a size of 896m$\times$896m, which is synthesized by $4\times4 = 16$ sub-regions with a size of $224\times224$. To avoid the edge artifacts resulted from convolution onto the edges when concatenating the sub-regions, we first extend the area of each sub-region to 448$\times$448 and then crop the 224$\times$224 region centrally to get the prediction of each sub-region. In such a way, the convolutional filters can also process the full local information on edges rather than the paddings. The visualization results are presented in Fig.~\ref{fig:exp_suppl}. To offer a more detailed view of the quality of the generated map, we visualize the whole test area generated by \model in the Porto dataset a representative. The result is shown in Fig.~\ref{fig:porto_test_vis}.

\section{Implementation Details}

\subsection{Data Preprocessing}
As introduced in the main text, we get the training sample by cropping an area of size 224$\times$224. In the following, we detail the procedure of constructing the training examples. 
An intuitive way to generate the training samples is cropping the whole city into disjoint $H\rm{m}\times W\rm{m}$ grids which are like meshes.
We do not adopt the mesh representation because it is not able to well preserve the information of grid boundaries. 
Instead, we adopt a different approach by randomly generating samples of size $H\rm{m}\times W\rm{m}$ from the whole city, excluding the testing and validation regions. The boundary of one sample could be in the center of another sample, which effectively addresses the limitation of mesh presentation. 
Our training set is generated in the same way, consisting of randomly generated training samples of fixed size. 
To guarantee fairness, we apply the same sampling strategy to all neural network-based baselines.  
We normalize the RGB channels of aerial image to $[0, 1]$ before feeding them to the network.
We use $224\times 224$ (\ie, $H=224$, $W=224$ and $\alpha$=1m/pixel) for implementation, which is the most common size for VGG-like models. However, the size is not fixed as the model uses a fully convolutional network and hence it can take in an image of any size as an input.

For trajectory feature extraction, as stated in the paper, we partition the region into many $\alpha\rm{m} \times \alpha\rm{m}$ grids to count the number of trajectory points falling in each grid. We have tried many different values for $\alpha$, \eg, 1m, 2m, 4m, 8m, and $\alpha=1$m turns out to be the best.   

As the training patch is randomly cropped from the whole area, it is hard to define what is ``an epoch'', as two training samples might have overlaps. As a solution, we introduce a number $N =\frac{S_{train}}{224 \times 224}$ with $S_{train}$ representing the area of the training region. $N$ actually defines the expected number of samples that could cover the entire training region, and an epoch refers to the scanning of the set of $N$ samples. 

\vspace{0.3in}
\subsection{Model Details}
\paragraph{Structure of U-Net.}
\label{sec:unet}
\model is designed based on U-Net. U-Net is a fully convolutional network for pixel-wise prediction through encoding and decoding stages. In the encoding stage, the spatial size of feature maps will be halved 4 times through max pooling, while the channels of feature maps will be doubled four times, which is similar to the front 8 layers in VGG network (type B) \cite{vgg}. The decoding stage up-samples the low-resolution encoded feature map to the original resolution prediction map through four $2\times2$ deconvolutions. One important feature of U-Net is the skip-connection from the feature map extracted in the encoding stage to the corresponding layers in the decoding stage to ensure the lower-level features, like shapes and edges, to be directly inferred in the decoding stage. Moreover, the gradient path will be also shortened via the short cut to relieve the gradient vanishing in deep neural networks. 

As the map extraction task (pixel-wise binary prediction) is expected to be no harder than the semantic segmentation task (pixel-wise multi-classification), we reduce the model parameters of the original U-Net for fast training and memory saving. In detail, we divide the output channels of all convolutional layers by 4. 
Note that we have conducted experiments to ensure that such a  modification will NOT harm the performance of \model. 
To guarantee a fair comparison, we divide the channels of convolutions of VGG-like models, \eg~SegNet, DeconvNet, and U-Net, by 4 too. 
The configurations of all layers of our modified U-Net are listed in Table~\ref{tab:unet_param}. 

\begin{table}[t]
	\caption{\label{tab:unet_param} Detailed configurations of the encoder and the decoder branches of our model. ``conv'' and ``deconv'' refer to the convolution layer and de-convolution layer respectively. Note that all ``conv'' layers are followed by a batch normalization layer and an ReLU activation. ``concat $x$'' denotes the concatenation of the outputs of the previous layer and the layer $x$.}
	\begin{center}
		\scriptsize
		\begin{tabular}{c|ccc|c}
			\hline
			name & kernel size & stride & pad & output size\\
			\hline
			image input & - & - & - & $224\times224\times3$\\
			traj input & - & - & - & $224\times224\times1$\\
			\hline
			\multicolumn{5}{c}{Encoder}\\
			\hline
			conv1-1 & $3 \times 3$ & 1 & 1 & $224\times224\times16$\\
			conv1-2 & $3 \times 3$ & 1 & 1 & $224\times224\times16$\\
			max-pool1 & $ 2 \time 2$ & 2 & 0 & $112\times112\times16$\\
			\hline
			conv2-1 & $3 \times 3$ & 1 & 1 & $112\times112\times32$\\
			conv2-2 & $3 \times 3$ & 1 & 1 & $112\times112\times32$\\
			max-pool2 & $ 2 \time 2$ & 2 & 0 & $56\times56\times32$\\
			\hline
			conv3-1 & $3 \times 3$ & 1 & 1 & $56\times56\times64$\\
			conv3-2 & $3 \times 3$ & 1 & 1 & $56\times56\times64$\\
			max-pool3 & $ 2 \time 2$ & 2 & 0 & $28\times28\times64$\\
			\hline
			conv4-1 & $3 \times 3$ & 1 & 1 & $28\times28\times128$\\
			conv4-2 & $3 \times 3$ & 1 & 1 & $28\times28\times128$\\
			max-pool4 & $ 2 \time 2$ & 2 & 0 & $14\times14\times128$\\			
			\hline
			conv5-1 & $3 \times 3$ & 1 & 1 & $14\times14\times256$\\
			conv5-2 & $3 \times 3$ & 1 & 1 & $14\times14\times256$\\
			\hline
			\multicolumn{5}{c}{Decoder}\\
			\hline
			deconv4-1 & $2 \times 2$ & 2 & 0 & $28\times28\times128$\\
			concat conv4-2 & - & - & - & $28\times28\times256$\\
			conv4-3 & $3 \times 3$ & 1 & 1 & $28\times28\times128$\\
			conv4-4 & $3 \times 3$ & 1 & 1 & $28\times28\times128$\\
			\hline
			deconv3-1 & $2 \times 2$ & 2 & 0 & $56\times56\times64$\\
			concat conv3-2 & - & - & - & $56\times56\times128$\\
			conv3-3 & $3 \times 3$ & 1 & 1 & $56\times56\times64$\\
			conv3-4 & $3 \times 3$ & 1 & 1 &$56\times56\times64$\\
			\hline
			deconv2-1 & $2 \times 2$ & 2 & 0 & $112\times112\times32$\\
			concat conv2-2 & - & - & - & $112\times112\times64$\\
			conv2-3 & $3 \times 3$ & 1 & 1 & $112\times112\times32$\\
			conv2-4 & $3 \times 3$ & 1 & 1 & $112\times112\times32$\\
			\hline
			deconv1-1 & $2 \times 2$ & 2 & 0 & $224\times224\times16$\\
			concat conv1-2 & - & - & - & $224\times224\times32$\\
			conv1-3 & $3 \times 3$ & 1 & 1 & $224\times224\times16$\\
			conv1-4 & $3 \times 3$ & 1 & 1 & $224\times224\times16$\\
			\hline
			\multicolumn{5}{c}{Predictor}\\
			\hline
			fc & $1\times 1$ & 1 & 0 & $224\times224\times2$\\
			softmax & - & - & - & $224\times224\times2$\\
			\hline
		\end{tabular}
	\end{center}    
\end{table}

\paragraph{Gated Fusion Module.}
Recall that the decoder phase of \model has 5 levels and it adopts the GFM at every level. Specifically, we assign individual weights i.e., $\lbrace a_I, a_T, \psi_{1\times 1}, \phi_{3\times3}^2\rbrace$ to the GFM of each level. The convolutions $a_I, a_T$ and $\phi_{3\times3}^2$ will not change the channel dimension; while the convolution $\psi_{1\times 1}$ will transfer the channel dimension to 2 to perform softmax normalization. The non-parametric $2\times$ upsampling operation $U_{2\times}(\cdot)$ is implemented by the nearest neighbor up-sampling strategy. Specifically, $\tilde{G}^{(0)}$ is set to $0$,  which serves as a non-informative prior.

\paragraph{Densely Supervised Refinement Decoding.}
Our dense supervision needs to provide labels in all scales, \ie, levels $1\sim 5$. We specify the generation of the labels in levels $1 \sim 4$. To get the ground truth of level $i$ with $i\in [1,4]$, we adopt the $2\times 2$ average pooling on the ground truth of level $i+1$. It means the label is a real value in the range of $(0, 1)$, which can be explained by the ratio of having roads in its place, and the cross-entropy loss is compatible with such labels.

\paragraph{Output Format.}
There are mainly two types of output format for map extraction task, \ie, the graph representation and the binary image representation. The former represents the map by a directional graph, usually used by the trajectory-based approaches (\citenp{cobweb}; \citenp{tc1}; \citenp{kde}; \citenp{trajsift}). It directly captures the topology of the road network but is hard to be optimized by end-to-end. The latter, usually used by aerial image-based approaches (\citenp{hinton_mapinference}; \citenp{casnet}; \citenp{robocodes}; \citenp{triseg}; \citenp{1ddecoder}; \citenp{dlinknet}; \citenp{stackedunet}), is easy to be transferred into a pixel-wise binary prediction task but loses the topological information. 

In this paper, we decide to use the binary image representation as the output format. This is because even if the graph representation is necessary for certain applications, we are able to generate the graph representation from the binary images, as approaches like (\citenp{deeproadmapper}; \citenp{roadtracer}) have been proposed recently to predict the graph representation based on post-processing a first-stage binary image prediction. Moreover, the recent DeepGlobe road extraction challenge also selected such image-based output format \footnote{https://competitions.codalab.org/competitions/18467}.

\section{Experiment Details}

\subsection{Dataset}
\label{sec:dataset}
\begin{figure*}[t]
	\centering
	\subfigure[Porto]
	{\label{fig:porto_dataset}\includegraphics[height=44mm]{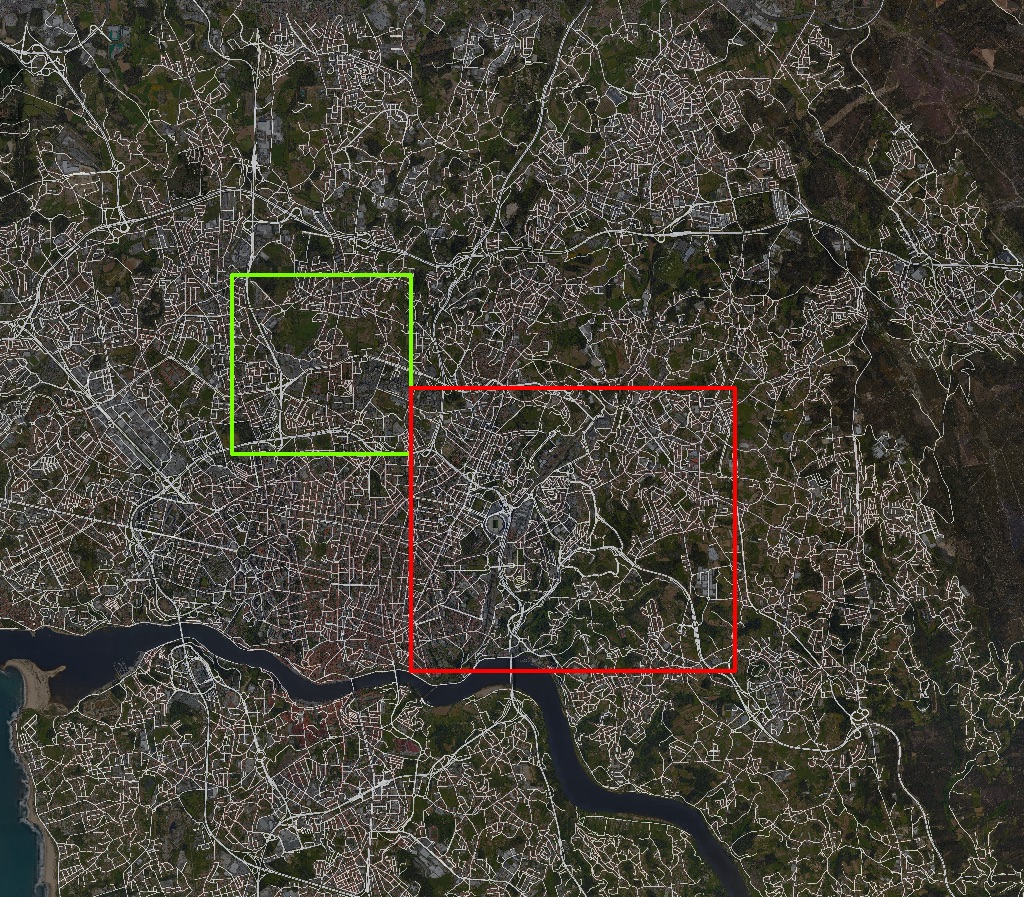}}
	\hspace{0.05in}
	\subfigure[Shanghai]
	{\label{fig:shanghai_dataset}\includegraphics[height=44mm]{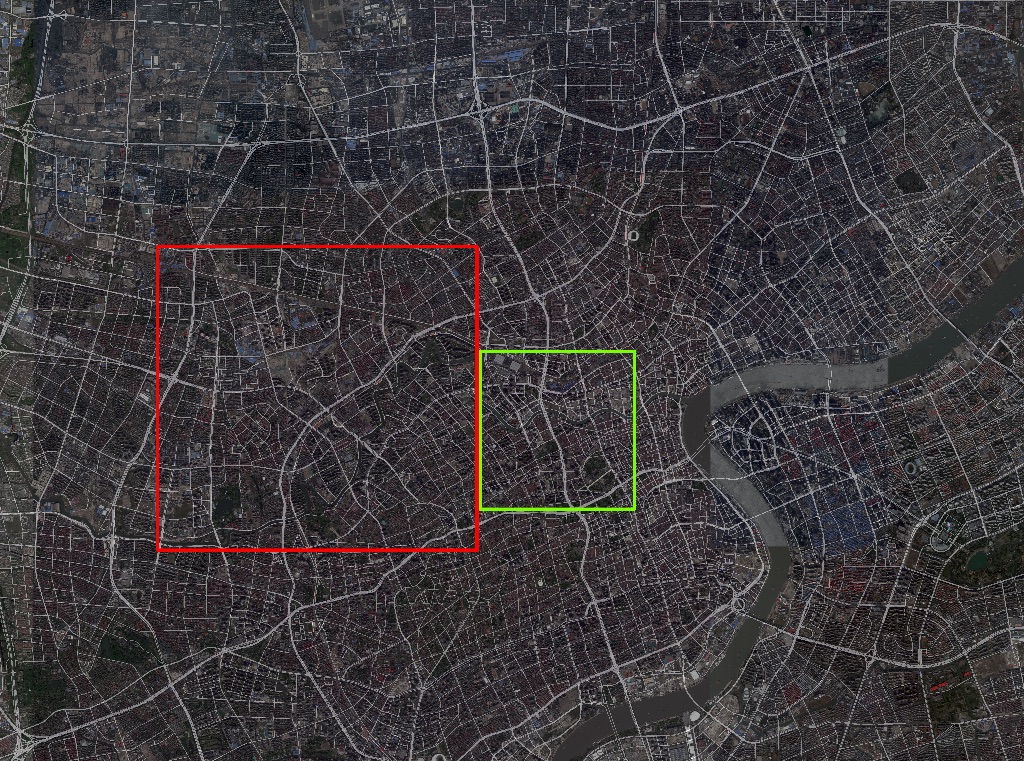}}
	\hspace{0.05in}
	\subfigure[Singapore]
	{\label{fig:singapore_dataset}\includegraphics[height=44mm]{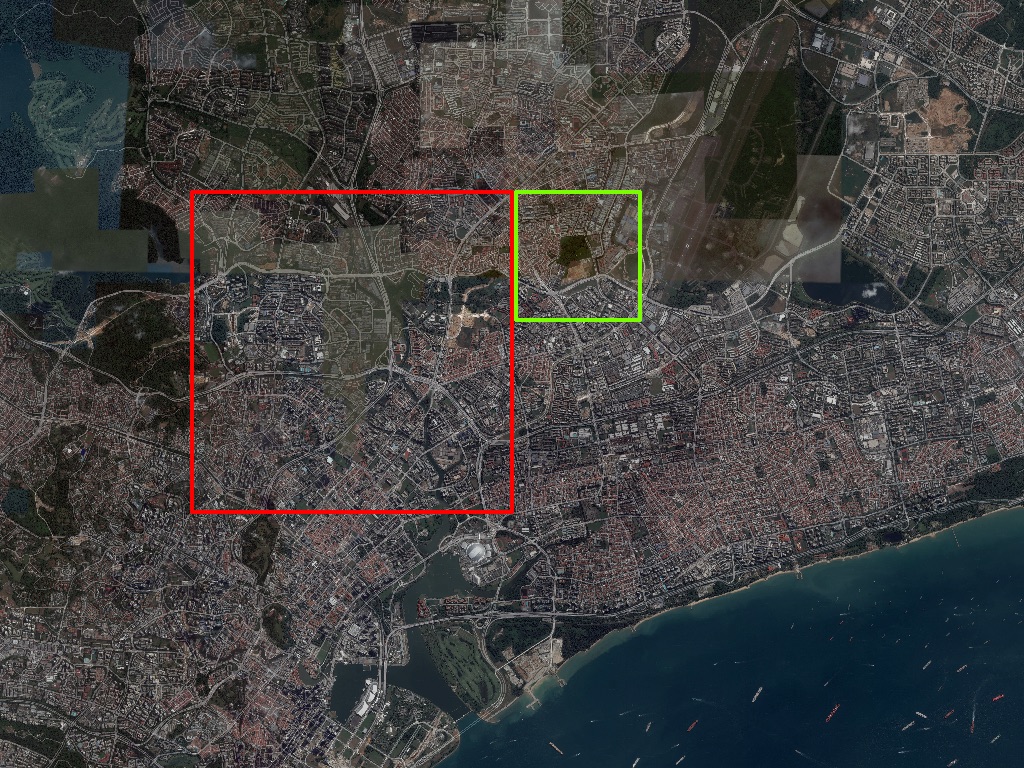}}
	\caption{Visualization of three datasets. The regions inside the green and red boxes refer to the validation area and the testing area respectively, while the remaining regions form the training set.}
	\label{fig:dataset}
\end{figure*}

Notice that the map extraction task studied in this paper leverages both aerial images and GPS trajectories, thus we are not able to directly adopt existing datasets for evaluation. 
We retrieve the aerial images from Google Map API (zoom=17) with the original resolution being roughly 1m/pixel, and we resize them to 1m/pixel for simplicity.
For GPS trajectory datasets, all trajectories record the journeys of taxis. The Porto dataset is an open-sourced dataset\footnote{Available at https://www.kaggle.com/c/pkdd-15-predict-taxi-service-trajectory-i/data}. It contains trajectories generated by 442 taxis from 2013 to 2014. 
The Shanghai dataset has trajectories generated by around 13,000 taxis for 3 days in 2015.
The Singapore dataset is generated by about 15,000 taxis for 60 days in 2012. 
Table~\ref{tab:dataset} lists the detailed statistics of all datasets.\footnote{https://github.com/algorithm-panda/DeepDualMapper}
Fig.~\ref{fig:dataset} visualizes the training, validation and testing areas.
The testing area (in red boxes) is selected such that it consists of both dense and sparse roads to test the performance of \model under different scenarios.

\paragraph{Ground Truth.}
Since the datasets used in this paper are not the public datasets whose roads have already been annotated, the ground truth is not available. To efficiently build the ground truth, we leverage the road networks from OpenStreetMap (www.openstreetmap.org). The latest version of maps are downloaded to ensure the completeness of the road networks. We exclude the roads belonging to the service type. Recall that in this paper, we regard the map extraction as a pixel-wise binary classification task. Thus, we need to transfer the road network into an image. In detail, we draw roads by lines at 10 pixel width to generate the image representation of the ground truth. The reason we select 10 pixel is that it represents 10 meters in the real world (as the aerial image has the resolution at 1m/pixel), which is the average road width. We have also tested the performances of \model using other widths while the setting of width does not affect the performance much. Notice that one may argue that our assumption of all the roads sharing the same width may be inconsistent with reality. 
We would like to highlight that it is not a concern as we have found ALL neural network-based models can learn such fixed-width automatically without much effort, even it is not consistent with the real width.


\begin{table}[htb]
	\caption{\label{tab:dataset} The statistics of three datasets under the areas shown in Fig.~\ref{fig:dataset}.}
   	\begin{center}
		\begin{tabular}{c|c|c|c}
			\hline
			Dataset & Porto & Shanghai & Singapore\\
			\hline
			Width (km) & 15.447 & 19.500 & 16.000 \\
			Height (km) & 13.538 & 14.500 & 12.000 \\
			Validation area percentage & 3.486\% & 3.183\% & 2.083\% \\
			Test area percentage & 9.998\% & 12.439\% & 13.021\% \\
			\# Trajectories (k) & 1,692 & 1,105 & 952 \\
			\# Trajectory pts (million) & 77 & 95 & 584 \\
			\# Trajectory pts per 1m$\times$1m grid & 0.367 & 0.335 & 3.043 \\
			\# Trajectory pts per valid 1m$\times$1m grid & 6.461 & 3.616 & 13.585 \\
			Trajectory sampling interval (s) & 15.012 & 9.391 & 25.135 \\
			\# Roads (k) & 43 & 21 & 37 \\
			\# Roads/$\rm{km}^2$ & 207.1 & 77.6 & 195.2 \\
			Original image resolution (m/px) & 0.90 & 1.02 & 1.19\\
			\hline
		\end{tabular}
	\end{center} 
\end{table}

\subsection{Metrics}
As introduced in the main text, we adopt IoU and F1-score as the evaluation metrics. In detail, we denote correctly predicted road pixels, \ie, true positive, as TP, correctly predicted non-road pixels, \ie, true negative, as TN, road pixels which are wrongly predicted as non-roads, \ie, false negative, as FN, and non-road pixels which are wrongly predicted as roads, \ie, false positive, as FP. The IoU is computed as $\frac{\#TP}{\#TP+\#FP+\#FN}$. Precision is computed as $\frac{TP}{TP+FP}$ and recall is computed as $\frac{TP}{TP+FN}$. F1-score, the harmonic mean of precision and recall, is computed as $\frac{2\times precision \times recall}{precision + recall}$. Note that the numbers of TP, FP, TN, and FN pixels are counted in the whole testing area.

\subsection{Training Details}
\label{sec:training}

This paper is to study the data fusion methods on aerial images and GPS trajectories for map extraction. Accordingly, we want to minimize the influence of other factors. To do so, we use the same data loader to provide training and testing samples for all neural network-based models. In detail, as introduced previously, in each step, we randomly crop a $224 \times 224$ area for training for all approaches. 
We also adopt the same learning rate (1e-4) and optimizer (Adam) on all the models. Notice that as the input features contain not only a pure RGB image but also 1-channel trajectory data, it is hard to leverage the pre-trained weight (e.g., via ImageNet) to initialize the kernel of convolutions. Fortunately, we found that all the models are able to converge fast even when trained from scratch, thus we randomly initialize all the parameters for all the models. 

There are some fluctuations in the performance of some approaches. As a solution, we train all the approaches 50 epochs (which is enough for them to converge) and report the average results of the last 10 epochs. 
Note that DeepRoadMapper and RoadTracer adopt their own segmentation network, and we observe that their performance converges slower than other baselines. Consequently, we train them 100 epochs and report the average performance of the last 10 epochs. Moreover, as the original DeepRoadMapper and RoadTracer both first produce the image representation of the map and then post-process it into the graph representation, we only include their segmentation networks to produce the map in image representation for a fair comparison.

We implement our code in PyTorch 0.4.1 under Ubuntu 16.04. The CPU is Intel Core i7-6850K with 128GB memories. All the models can fit in an Nvidia GTX 1080Ti GPU having 11GB memories. Our model takes about 7.6ms to infer a patch in $224 \times 224$ (with batch size 16).

\end{document}